\documentclass[a4paper]{cas-dc}

\usepackage[authoryear]{natbib}
\usepackage{amsmath}
\usepackage{amssymb}
\usepackage{amsfonts}
\usepackage{hyphenat} 
\usepackage{graphicx}
\usepackage{subfigure}
\usepackage{caption}
\usepackage{subcaption}
\usepackage{bbding} 
\usepackage{multirow}
\usepackage{multicol}

\usepackage{threeparttable}
\usepackage{color}
\usepackage{xcolor}
\usepackage{colortbl}

\newcommand{\ie}{\textit{i}.\textit{e}., }
\newcommand{\eg}{\textit{e}.\textit{g}.}
\newcommand{\etc}{\textit{etc}. }

\definecolor{colorFst}{HTML}{bde6cd}       
\definecolor{colorSnd}{HTML}{e4eebc}       
\definecolor{colorTrd}{HTML}{fff8c5}       

\newcommand{\fs}{\cellcolor{colorFst}\bf}   
\newcommand{\nd}{\cellcolor{colorSnd}}      
\newcommand{\rd}{\cellcolor{colorTrd}}      

\def\tsc#1{\csdef{#1}{\textsc{\lowercase{#1}}\xspace}}
\tsc{WGM}
\tsc{QE}
\tsc{EP}
\tsc{PMS}
\tsc{BEC}
\tsc{DE}

\begin{document}
\let\WriteBookmarks\relax
\def\floatpagepagefraction{1}
\def\textpagefraction{.001}

\shorttitle{GOOD: Towards Domain Generalized Oriented Object Detection}

\shortauthors{Bi et~al.}

\title [mode = title]{
GOOD: Towards Domain Generalized Oriented Object Detection
}



%
\author[1,2]{Qi Bi}
\author[1]{Beichen Zhou}
\author[1]{Jingjun Yi}
\author[3]{Wei Ji}
\author[4]{Haolan Zhan}
\author[1]{Gui-Song Xia}
\cormark[1]




\affiliation[1]{organization={School of Computer Science, Wuhan University},
    city={Wuhan},
    postcode={430072}, 
    country={China}}
\affiliation[2]{organization={Faculty of Science, University of Amsterdam},
    city={Amsterdam},
    postcode={1098XH}, 
    country={The Netherlands}}
\affiliation[3]{organization={School of Medicine, Yale University},
    city={New Haven},
    postcode={CT 06520}, 
    country={United States}}
\affiliation[4]{organization={Faculty of Information Technology, Monash University},
    city={Melbourne},
    postcode={VIC 3800}, 
    country={Australia}}

\cortext[cor1]{Corresponding authors.}



\begin{abstract}
Oriented object detection has been rapidly developed in the past few years, but most of these methods assume the training and testing images are under the same statistical distribution, which is far from reality.
In this paper, we propose the task of domain generalized oriented object detection, which intends to explore the generalization of oriented object detectors on arbitrary unseen target domains.
Learning domain generalized oriented object detectors is particularly challenging, as the cross-domain style variation not only negatively impacts the content representation, but also leads to unreliable orientation predictions.
To address these challenges, we propose a generalized oriented object detector (GOOD).
After style hallucination by the emerging contrastive language-image pre-training (CLIP),
it consists of two key components, namely, rotation-aware content consistency learning (RAC) and style consistency learning (SEC).
The proposed RAC allows the oriented object detector to learn stable orientation representation from style-diversified
samples.
The proposed SEC further stabilizes the generalization ability of content representation from different image styles.
Notably, both learning objectives are simple, straight-forward and easy-to-implement. 
Extensive experiments on multiple cross-domain settings show the state-of-the-art performance of GOOD.
Source code will be publicly available.
\end{abstract}


\begin{highlights}
    \item We make an early exploration to the task of learning domain generalized oriented object detectors. The evaluation protocols of this task are accordingly established.
     \item A generalized oriented object detector, dubbed as GOOD, is proposed. With CLIP-driven style hallucination, it consists of a rotation-aware consistency (RAC) and a style consistency (SEC) learning to represent generalized oriented objects. 
     \item The proposed RAC is innovative to constrain the both horizontal and rotated regions of interest (HRoIs and RRoIs). Technically, the pre-
     and post- hallucinated HRoIs and RRoIs that have the most similar orientation information is aligned, while the rest are pushed as apart as possible.
     \item The proposed GOOD significantly outperforms existing state-of-the-art oriented object detectors on a variety of generalization settings, by up to 3.29\% mAP on DOTA and 3.08\% on SODA. 
\end{highlights}

\maketitle

\section{Introduction}
\label{sec1}

\subsection{Problem Statement}

Oriented object detection occupies a unique intersection between computer vision and remote sensing \cite{xia2018dota,ding2021object,yang2021r3det,yang2021learning}, involving the prediction of oriented bounding boxes for objects in aerial images. 
Given the bird's-eye view perspective of an image sensor \cite{yao2022r2ipoints,Yang2023SAGN,bi2021local,bi2022all,bi2020multiple,zhou2021differential}, objects in aerial images pose arbitrary orientations. 
Therefore, in addition to the conventional task of accurately identifying and localizing objects \cite{tian2019fcos,lin2017focal,li2022exploring}, an oriented object detector must also provide precise predictions of an object's rotation angle \cite{qian2021learning,yang2020arbitrary,yang2021dense,yang2022kfiou}.

\begin{figure}[!t]
  \centering
   \includegraphics[width=0.49\textwidth]{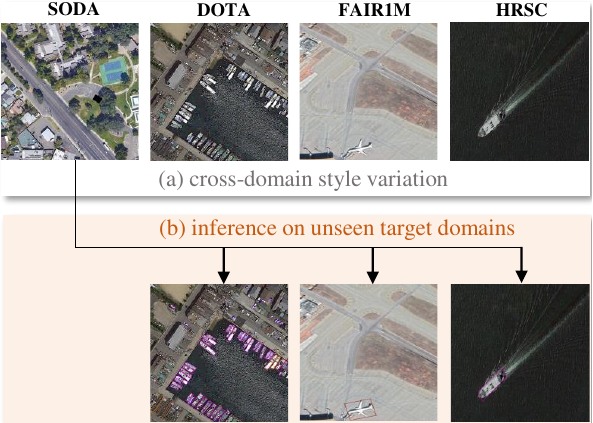} 
   \caption{(a) Domain generalized oriented object detection aims to learn an oriented object detector from only source domain that can be well generalized to (b) arbitrary unseen target domains. }
   \label{motivation}
\end{figure}

Existing oriented object detectors typically rely on the assumption that the source domain during training and the target domain during inference share the same statistical distribution. 
However, this assumption often diverges from reality. 
As shown in Fig.~\ref{motivation}a, aerial images from different domains have great style variation, caused by illumination, imaging sensor, geo-location, landscape and \etc \cite{iizuka2023frequency,zhu2023style}.
A pre-trained oriented object detector on WorldView-2 images from American landscape (\eg, SODA \cite{Cheng2023SODA} as source domain) need to conduct inference on numerous unseen target domains, \eg, Gaofen-2 images from Chinese landscape in DOTA \cite{xia2018dota} domain.

\begin{figure}[!t]
  \centering
   \includegraphics[width=0.49\textwidth]{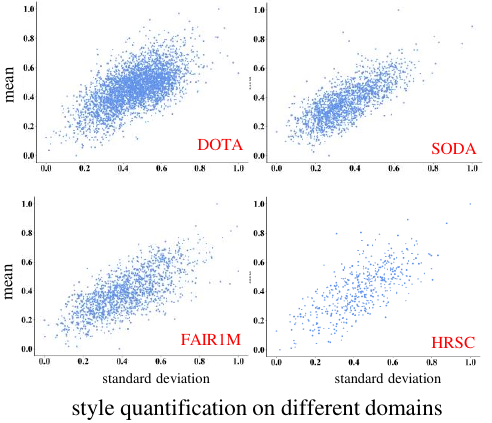} 
   \caption{Quantifying the styles from different aerial image domains. The per-sample style is quantified by two variables, namely, mean and standard deviation \cite{huang2017arbitrary}. Mean demonstrates how the samples are spread out in the feature space, whereas standard deviation demonstrates how the per-sample feature varies. For each aerial image domain, the per-sample mean and standard deviation is computed and visualized. Styles from different domains show a dramatic discrepancy.}
   \label{style}
\end{figure}

To better understand the domain shift caused by styles, we quantify the style of aerial images from different domain following the existing definition of styles \cite{gatys2016image,huang2017arbitrary}, where the channel-wise mean and standard deviation is used.
We compute these two metrics of each sample in each domain, and visualize the style distribution of these domains.
As shown in Fig.~\ref{style}, the style distribution of each domain is quite different, which accounts for the domain gap.
Developing a domain generalized oriented object detector 
enables the inference across diverse and previously unobserved target domains (Fig.~\ref{motivation}b). 
Regrettably, as far as our knowledge extends, this area remains largely unexplored.

\subsection{Motivation \& Objective}

In this paper, we make an early exploration to learn domain generalized oriented object detectors.
The great style variation between the source domain and unseen target domains is usually reflected from the image appearances that have different color distribution (\eg, intensity, contrast) \cite{lee2022bridging,Robust2021,nam2021reducing,bi2024CMFormer}, which are the input of an oriented object detector pre-trained on a certain source domain.
The different color distribution makes the pre-trained oriented object detector difficult to perceive the activation patterns and to build the feature representation as learned on the source domain \cite{huang2023reciprocal,Robust2021}.
The shift of activation patterns on the target domains inevitably poses two key challenges for oriented object detectors.

\begin{figure}[!t]
  \centering
   \includegraphics[width=0.49\textwidth]{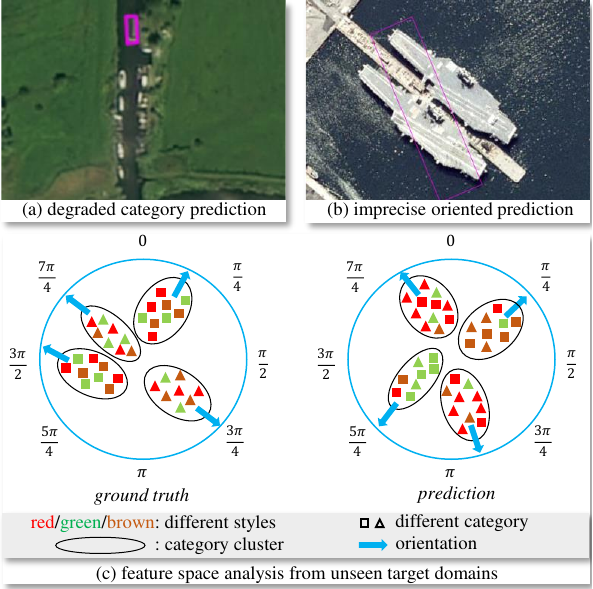} 
   \caption{
   Two key challenges when an oriented object detector generalizes to unseen target domains. (a) \textbf{degraded category prediction}: multiple ships are not detected. In the feature space, samples from each category are not properly clustered due to cross-domain style variation; (b) \textbf{imprecise oriented prediction}: the rotation angle does not align with the object orientation, which is reflected in both image space and blue arrows in feature space.
   }
   \label{challenge}
\end{figure}

First of all, the shift of activation patterns accumulates from shallow to deep of an oriented object detector, and inevitably confuses and degrades the high-level representation, \eg, region of interests (RoIs).
Consequently, as illustrated in Fig.~\ref{challenge}a,
the category-level representation, also known as a kind of content information, can be less reliable for a detector to precisely identify an object.
A same object category from different styles can be difficult to cluster into the same category.

Secondly, such statistic variation from the target domain poses a unique challenge to predict rotated bounding boxes than conventional horizontal bounding boxes, \ie less representative orientation information (in Fig.~\ref{challenge}b).
This is because the orientation is usually regressed from the RoIs. As the RoIs degrade on the target domains, the orientation inevitably becomes less precise. As illustrated in Fig.~\ref{challenge}c, the orientation of the prediction can be quite different from the ground truth.

We propose a domain generalized oriented object detector, dubbed as GOOD, to address the above challenges.
First, we aim to allow the GOOD to perceive more diverse styles when training on the source domain.
The emerging contrastive language-image pre-training (CLIP) \cite{radford2021learning} rests in abundant styles from web-wide images, and provides us a feasible solution.
Afterwards, the proposed GOOD consists of two key components, namely, rotation-aware content consistency learning (RAC) and style consistency learning (SEC).
The proposed RAC constrains both horizontal region of interests (HRoIs) and rotated region of interests (RRoIs) between the original images and style-hallucinated images, so that both the orientation and content representation is more robust to the cross-domain style variation.
On the other hand, the proposed SEC poses a category-wise constraint between the original and style-hallucinated images, which allows the style-hallucinated images to perceive the same content.
In this way, the classification head yields more discriminative category prediction on the unseen target domains.

Notably, although both RAC and SEC are simple, straight-forward and easy-to-implement, they still advance the contrastive learning and consistency learning in multiple predominant perspectives. In contrast to the generic contrastive learning works, RAC focuses on the simultaneous constraint of both horizontal and rotated regions of interests (HRoIs and RRoIs). For both HRoIs and RRoIs, the pre- and post- hallucinated regions of interests that have the most similar orientation information are aligned, while the rest are pushed as apart as possible. It allows the orientation information on unseen target domains to be more stable. Besides, though the Jensen-Shannon Divergence is commonly used for consistency constraint, the proposed SEC is innovative in two-fold. Firstly, it brings up a new style hallucination method from per-layer vision-language foundation model (VFM) features. 
In contrast, existing style hallucination methods usually only hallucinate the last-layer feature from pre-trained ImageNet encoder. Our method allows a more fine-grained and progressive hallucination from CLIP encoder with more diverse styles. Besides, hallucinating from CLIP features has also not been explored before.

Last, but not least, it is highly important to establish the evaluation protocols for domain generalized oriented object detection, which provides a solid baseline for future study.
Extensive domain generalization settings on existing oriented object detection benchmarks (\ie, FAIR1M \cite{sun2022fair1m}, DOTA \cite{xia2018dota}, SODA \cite{Cheng2023SODA}, HRSC \cite{liu2016ship}) have been designed to benchmark domain generalized oriented object detection.
Following prior driving-scene object detection works \cite{wu2022single,lin2021domain,vidit2023clip}, the common object categories among each dataset are used for experiments.
Specifically, four cross-domain settings, namely, FAIR1M-10 classes, SODA-7 classes, DOTA-4 classes and HRSC-ship, are involved.
The proposed GOOD shows substantial improvement on the multiple baselines on all the generalization settings.

\subsection{Contribution}

Our contribution can be summarized as below.
\begin{itemize}
\item We make an early exploration to the task of learning domain generalized oriented object detectors. The evaluation protocols of this task are accordingly established.
\item A generalized oriented object detector, dubbed as GOOD, is proposed. With CLIP-driven style hallucination, it consists of a rotation-aware consistency (RAC) and a style consistency (SEC) learning to represent generalized oriented objects. 
\item The proposed RAC is innovative to constrain the both horizontal and rotated regions of interest (HRoIs and RRoIs). Technically, the pre-
and post- hallucinated HRoIs and RRoIs that have the most similar orientation information is aligned, while the rest are pushed as apart as possible.
\item The proposed GOOD significantly outperforms existing state-of-the-art oriented object detectors on a variety of generalization settings, by up to 3.29\% mAP on DOTA and 3.08\% on SODA. 
\end{itemize}

The remainder of this paper is organized as follows. In Section~\ref{sec2}, related work is provided. In Section~\ref{sec3}, the proposed method is demonstrated. In Section~\ref{sec4}, we report and discuss the experiments on three aerial image scene classification benchmarks. Finally in Section~\ref{sec5}, the conclusion is drawn.

\section{Related Work}
\label{sec2}

\subsection{Oriented Object Detection}

Existing works can be categorized into two types, namely, two-stage and one-stage.
Horizontal bounding box \cite{deng2018multi,liu2021abnet,zhu2023knowledge,tan2023wsodet} is less precise to depict the spatial position and the orientation of an object.
Early rotated bounding box supervised works usually modified the anchor related modules \cite{tian2019fcos,lin2017focal} for oriented objects. 

Typical works of two-stage oriented object detectors include RoI Transformer \cite{ding2019learning}, ReDet \cite{han2021redet}, Oriented R-CNN \cite{xie2021oriented}, JDet \cite{xiao2024theoretically} and \etc. 
One-stage oriented object detectors can be further categorized into three types.
The first type (\eg, R3Det \cite{yang2021r3det}, $S^2$A-Net \cite{han2021align}, DCFL \cite{xu2023dynamic}) focus on feature alignment between the regression branch and classification branch.
The second type focus on angle loss regression \cite{qian2021learning,yang2020arbitrary,yang2021dense,yang2021learning,yang2022kfiou,yu2023phase,xu2024rethinking}.
The third type follows the RepPoint paradigm, which leverages the point prediction to bound an oriented object \cite{hou2023g,li2022oriented}. 

More recently, some recent works jointly leverage the advantages of both the first paradigm and the second paradigm. For example, \cite{qiao2023novel} proposes an adaptive angle optimization head on top of the conventional regional proposal branch. \cite{ming2023task} proposes a framework that jointly improves the image representation from the region proposal network, the orientation estimation branch and the label assignment branch.

However, these oriented object detectors have an obvious weakness that they assume the training and inference objects are under the same distribution.
When there is large feature distribution shift caused by style variation between the source and target domains, both the rotation angle prediction and the category prediction can suffer severe degradation.
Several recent works tackle this challenge by domain adaptation, which still need both source domain and target domain when training \cite{liu2024clip,liu2024source}. Ideally, the oriented object detector, only trained on a source domain, is supposed to be well generalized to arbitrary unseen target domains.

To learn domain generalized oriented object detectors, we choose the two-stage paradigm not only because it provides a more precise description between orientation and region of interests, but also it provides more access to enrich the style and constrain the content consistency from shallow to deep.

\subsection{Domain Generalized Object Detection}

Domain generalized object detection is more challenging than domain adaptive object detection \cite{li2022cross,li2022source,zhao2022task}. 
It aims to learn an object detector from only the source domain data that can be generalized to unseen target domains \cite{lin2021domain}.
More recent works focus on learning domain generalized object detectors on scenarios such as driving scenes \cite{wu2022single,lin2021domain,vidit2023clip}, 3D point cloud \cite{lehner20223d,wang2023towards}, UVA image \cite{wang2023generalized} and \etc. 
On the other hand, domain generalization has received increasing attention.
Multiple advanced machine learning methods (\eg, entropy regularization \cite{zhao2020domain}, common-specific low-rank decomposition \cite{piratla2020efficient}, casual matching \cite{mahajan2021domain}, extrinsic-intrinsic interaction \cite{wang2020learning}, balance invariance \cite{chattopadhyay2020learning}, batch normalization embeddings \cite{segu2023batch}) have been proposed to enhance the generalization ability.
An extensive survey can refer to \cite{zhou2022domain}.
Meanwhile, some recent works studied domain generalization under the single source domain setting \cite{qiao2020learning,peng2022out,zhou2020learning,bi2024fog,yi2024learning,yi2024hallucinated} or unsupervised setting \cite{harary2022unsupervised,hu2022feature}. 

However, \textit{to the best of our knowledge}, none of these works focus on learning oriented object detectors. 
They predict horizontal bounding boxes \cite{wu2022single,liu2023periodically,vidit2023clip}, but are not applicable to regress the orientation angle.
In other words, it is difficult to fit them into the oriented object detection task. 

\subsection{Vision-language Model Driven Detection}

Contrastive language-image pre-training (CLIP) \cite{radford2021learning} has recently become a domaint visual representation learning trend, which utilizes the image-text pairs to learn visual-semantic representation by pre-training \cite{radford2021learning,desai2021virtex,engilberge2018finding}.
For object detection community, CLIP has significantly improved the visual representation for open vocabulary object detectors \cite{zareian2021openvocabulary,gu2022openvocabulary,bangalath2022bridging,liunian2022grounded,zhang2022glipv2}, which involve detecting objects based on arbitrary textual descriptions. 

More recently, Single-DGOD \cite{vidit2023clip} utilizes CLIP for domain generalized detectors on driving scenes.
The general idea is that, CLIP is pre-trained on large-scale image-text pairs, which contains much more abundant style information than naive augmentation or ImageNet pre-trained model. It provides the possibility to enrich the image styles.
In \cite{vidit2023clip}, although the proposed DomainGen does not rely on the CLIP image encoder, it extracts a more generalized feature embedding from the CLIP text encoder, given the inputted scene description and domain description. It also relies on the stronger generalization ability of the large-scale language-image pairs for pre-training. Besides, it is not able to handle the varied orientation in the oriented object detection task.
Meanwhile, style augmentation is a common and effective path to enhance the generalization ability to unseen domains. Some recent works have leveraged CLIP to enhance the style diversity \cite{cho2023promptstyler,patashnik2021styleclip,kwon2022clipstyler}, which motivate us to implement style hallucination by CLIP on source domain images.

\section{Methodology}
\label{sec3}

\begin{figure*}[!t]
  \centering
   \includegraphics[width=1.0\textwidth]{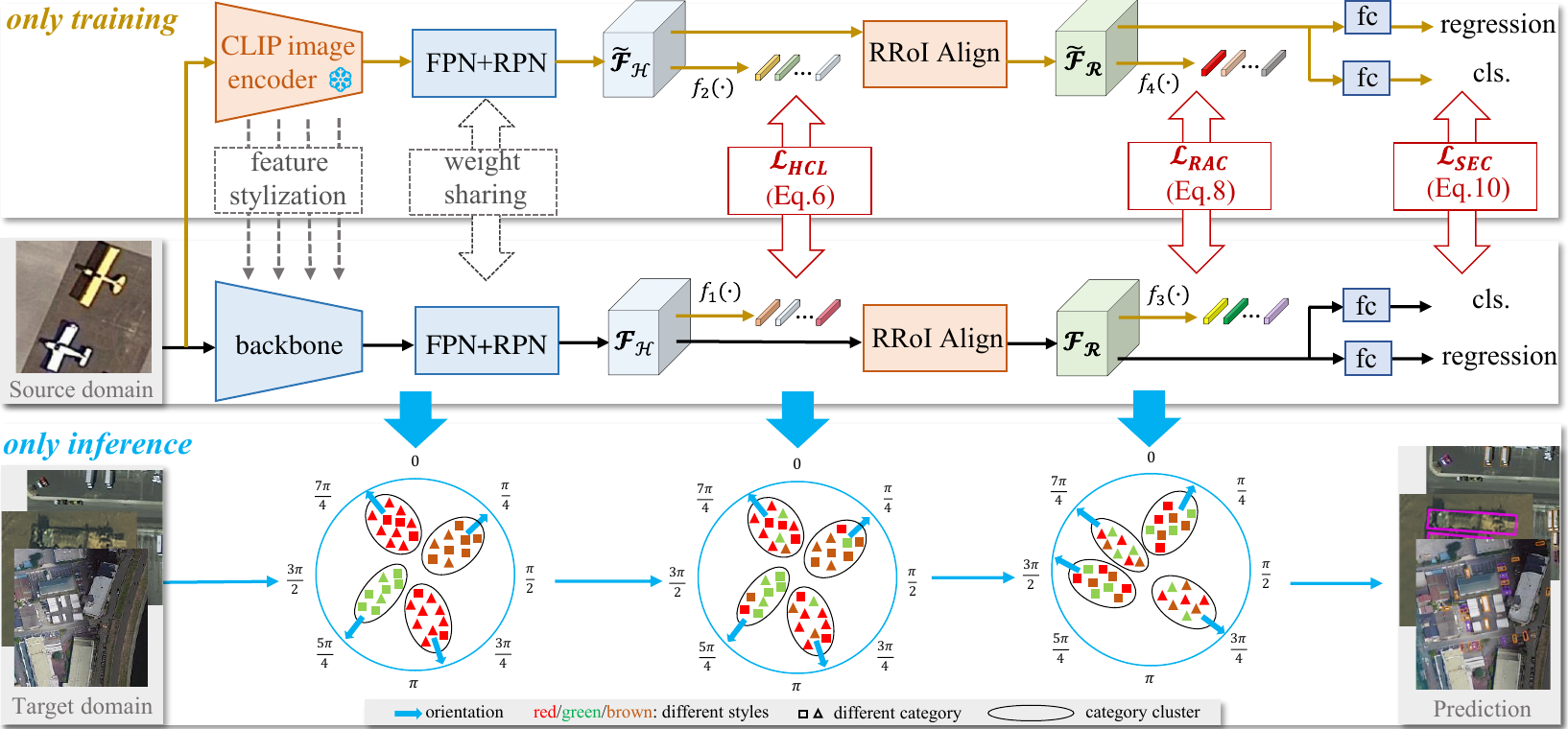} 
\caption{Framework overview of the proposed GOOD. After CLIP-driven style hallucination (Sec.~\ref{sec3.2}), two key components, namely, rotation-aware consistency learning (RAC, in Sec.~\ref{sec3.3}) and style consistency learning (SEC, in Sec.~\ref{sec3.4}), are involved. RAC implements content consistency on both horizontal and rotated region of interests (HRoI and RRoI), as presented in Eq.~\ref{HRoICL} and Eq.~\ref{RRoICL}, respectively. SEC implements style consistency on category-wise representation from both original and style-hallucinated images (in Eq.~\ref{JSDloss}). }
   \label{framework}
\end{figure*}

\subsection{Problem Setup \& Framework Overview}
\label{sec3.1}

Assume we have a source domain $\mathcal{S}$ and multiple unseen target domains $\mathcal{T}_1$, $\cdots$, $\mathcal{T}_k$, $\cdots$.
In general, the source domain and unseen target domains share the same label space but are under different distributions.
For domain $\mathcal{S}$, the joint image and label pair is $\{\mathbf{I}_n^{(\mathcal{S})}, (x_n^{(\mathcal{S})}, y_n^{(\mathcal{S})}, w_n^{(\mathcal{S})}, h_n^{(\mathcal{S})}, \theta_n^{(\mathcal{S})})\}_{n=1}^{N_s}$, where $n=1, 2, \cdots, N_s$.
Here $x_n^{(\mathcal{S})}, y_n^{(\mathcal{S})}, w_n^{(\mathcal{S})}, h_n^{(\mathcal{S})}, \theta_n^{(\mathcal{S})}$ denote the upper left x-coordinate, upper left y-coordinate, width, height and oriented angle of an oriented region of interest (RRoI), which jointly describe its 2D location in image $\mathbf{I}_n^{(\mathcal{S})}$.
$N_s$ denotes the sample number in domain $\mathcal{S}$.

The learning objective is to learn a oriented object detector $\phi: \mathbf{I} \rightarrow (x, y, w, h, \theta)$ only using the samples from source domain $\mathcal{S}$, and the model $\phi$ is supposed to generalize well on a unseen target domain $\mathcal{T}_{k}=\{(\mathbf{I}_n^{(k)})\}_{n=1}^{N_k}$.

Fig.~\ref{framework}~gives an overview of the proposed generalized oriented object detector, dubbed as GOOD.
After the feature extraction from a backbone, it has two key components, namely, Rotation-aware Consistency Learning (RAC) and Style Consistency Learning (SEC), which are used to decouple the negative impact of different styles on rotation prediction and category prediction, respectively.
Overall, it is backbone independent, and can be deployed into off-the-shelf two-stage oriented object detectors that have a step of rotation region of interest (RROI) representation and regression \& classification prediction.

\subsection{CLIP-driven Style Hallucination}
\label{sec3.2}

When training on the source domain, a domain generalized oriented object detector is supposed to encounter as many styles as possible. 
CLIP \cite{radford2021learning} learns image representation from webwide image-text pairs, which have seen extensive styles in the training stage.  
The abundant style information could benefit an oriented object detector when generalized to unseen target domains that have unseen styles.
To this end, we propose a novel CLIP-driven style hallucination scheme to enrich the style diversity to learn a generalized oriented object detector.

As an object detector usually leverages features from each stage of a backbone to represent the region of interest, it is necessary to augment the style variation on the features from each stage. 
Specifically, 
let $f_{\mathcal{I}}$ and $f$ denote the CLIP image encoder and the backbone of an object detector, respectively. 
The CLIP image encoder is ResNet-50.
From shallower to deeper, assume $\mathbf{F}_{\mathcal{I}}^{i} \in \mathbb{R}^{W_i \times H_i \times C_i}$ and $\mathbf{F}^{i}\in \mathbb{R}^{W_i \times H_i \times C_i}$ denote the feature representation from the $i^{th}$ block ($i=1, \cdots, 4$) of the CLIP image encoder and the detector backbone, respectively, given by
\begin{equation} 
\label{eq:multiscaleC}
(\mathbf{F}_{\mathcal{I}}^{1},\mathbf{F}_{\mathcal{I}}^{2},\mathbf{F}_{\mathcal{I}}^{3},\mathbf{F}_{\mathcal{I}}^{4})=f_{\mathcal{I}}(\mathbf{I}),
\end{equation}
\begin{equation} 
\label{eq:multiscaleI}
(\mathbf{F}^{1},\mathbf{F}^{2},\mathbf{F}^{3},\mathbf{F}^{4})=f(\mathbf{I}).
\end{equation}

Take the feature representation $\mathbf{F}_{\mathcal{I}}^{i}$ from CLIP image encoder as an example.
Following existing style-transfer methods \cite{huang2017arbitrary,chen2021hinet}, the channel-wise normalization mean $\boldsymbol{\mu}_{\mathcal{I}}^{i}$ and standard deviation $\boldsymbol{\sigma}_{\mathcal{I}}^{i}$ carries the style of an image $\mathbf{I}$, computed as 
\begin{equation} 
\label{mean}
\boldsymbol \mu_{\mathcal{I}}^{i} =  \frac{1}{H_i W_i} \mathbf{F}_{\mathcal{I}}^{i},
\end{equation}
\begin{equation} 
\label{covariance}
\boldsymbol{\sigma}_{\mathcal{I}}^{i} =  \sqrt{ \frac{1}{H_i W_i} \sum_{h \in H_i, w \in W_i} (\mathbf{F}_{\mathcal{I}}^{i} - \boldsymbol \mu_{\mathcal{I}}^{i})^2}.
\end{equation}

Similarly, let $\boldsymbol{\mu}^i$ and $\boldsymbol{\sigma}^i$ denote the mean and standard deviation of the feature representation $\mathbf{F}^{i}$ from the detector backbone $f$.
Then, the style-diversified counterpart  $\mathbf{\widetilde{F}}^{i}$ from $\mathbf{F}^{i}$, which carries the style information from CLIP encoder, can be readily used by AdaIN \cite{huang2017arbitrary}, given by
\begin{equation} 
\label{decoloss}
\mathbf{\widetilde{F}}^{i} =  \boldsymbol{\sigma}_{\mathcal{I}}^{i} \cdot \frac{\mathbf{F}^{i}-\boldsymbol{\mu}^{i}}{\boldsymbol{\sigma}^{i}}+ \boldsymbol \mu_{\mathcal{I}}^{i}.
\end{equation}

After that, both the feature representation from the backbone $\mathbf{F}^{i}$ and after style hallucination $\mathbf{\widetilde{F}}^{i}$ ($i=1, \cdots, 4$) are processed by a feature pyramid network (FPN) to learn the fused feature representation before and after style hallucination, which are denoted as $\mathbf{F}$ and $\mathbf{\widetilde{F}}$, respectively. 

\subsection{Rotation-aware Consistency Learning}
\label{sec3.3}

Even if the hallucinated style can be assumed, learning a domain generalized oriented object detector is still particularly challenging, as the orientation information of the content under varied styles need to keep consistent.
Thus, we propose a rotation-aware consistency (RAC) learning to handle this issue.
As the activation patterns inevitably shift on the target domains and yield less discriminative RoIs (discussed in Sec.~\ref{sec1}), we consider to pose RAC between the RoIs from the original samples and the RoIs from the style-hallucinated samples.

Given a typical two-stage oriented object detector, after extracting feature representation $\mathbf{F}^4$ from the backbone, the regional proposal network (RPN) generates $n$ horizontal region of interests (HRoIs) $\{\mathcal{H}_i\}$, where $i=1,\cdots, n$.
Generally, $\mathcal{H}_i$ is under the format of $\{\overline{x}, \overline{y}, \overline{w}, \overline{h}\}$, which describes the 2D location of a RoI.
Then, $\{\mathcal{H}_i\}$ need to be mapped to rotated RoIs (RRoIs) $\{\mathcal{R}_i\}$ to represent the orientation information, where $\mathcal{R}_i$ is under the format of $(x, y, w, h, \theta)$.

An ideal domain generalized oriented object detector need to stabilize both rotation and content representation for both $\{\mathcal{H}_i\}$ and $\{\mathcal{R}_i\}$, despite the great style variation for samples from different domains.

The HRoIs' feature representation of the original image $\mathbf{F}_{\mathcal{H}}$ and the style hallucinated image $\mathbf{\widetilde{F}}_{\mathcal{H}}$ are learned from $\mathbf{F}$ and $\mathbf{\widetilde{F}}$ by region proposal and feature alignment, respectively. 
To maintain the content consistency before and after style hallucination, both HRoIs' representations $\mathbf{F}_{\mathcal{H}}$ and $\mathbf{\widetilde{F}}_{\mathcal{H}}$ are projected into the embeddings $\mathbf{z}_{\mathcal{H}} \in \mathbb{R}^{d \times n}$ and $\mathbf{\widetilde{z}}_{\mathcal{H}}  \in \mathbb{R}^{d \times n}$
by two MLPs $f_1(\cdot)$ and $f_2(\cdot)$, respectively.
Here $d$ refers to the dimension size of each embedding. 

Let $\mathbf{z}=[\mathbf{z}_{\mathcal{H}},  \mathbf{\widetilde{z}}_{\mathcal{H}}] \in \mathbb{R}^{d \times 2n}$, and let $z_j, z_k \in \mathbf{z}$, where $z_j, z_k \in \mathbb{R}^{d \times 1}$, $j, k=1, \cdots, 2n$ and $j \neq k$.
Assume $z_{j,+}$ and $z_j$ construct a positive pair, the content consistency learning objective for HRoIs (denoted as $\mathcal{L}_{HCL}$) can be computed as
\begin{equation} 
\label{HRoICL}
\mathcal{L}_{HCL} =  -\sum_{j=1}^{2n} \sigma(\mathcal{H}_j) \cdot {\rm log} \frac{exp(z_j \cdot z_{j,+} / \tau)}{ \sum_{j=1, j \neq k}^{2n} exp(z_j \cdot z_{k} / \tau)},
\end{equation}
where $\tau$ is a temperature hyper-parameter, and is by default set to be 1.0.
$\sigma(\cdot)$ is a binary function determined by the mean intersection over union (mIoU) value between a region of interest (\eg, $\mathcal{H}_j$) and ground truth (denoted as $\rm{gt}$), given by
\begin{equation} 
\label{sigmamiou}
\sigma(\mathcal{H}_j)=
\begin{cases}
1 & \text{if mIoU$(\mathcal{H}_j,\rm{gt})\geq$0.5 }\\
0 & \text{else}
\end{cases}.
\end{equation} 

The feature representation $\mathbf{F}_{\mathcal{R}}$ of RRoIs $\{\mathcal{R}_i\}$ is computed from the feature representation $\mathbf{F}_{\mathcal{H}}$ of HRoIs $\{\mathcal{H}_i\}$, usually by the RRoI alignment transformation \cite{ding2019learning}. 
Let $\mathbf{\widetilde{F}}_{\mathcal{R}}$ denote the feature representation of style hallucinated RRoIs $\{\mathcal{\widetilde{R}}_{i}\}$. 
Both $\mathbf{F}_{\mathcal{R}}$ and $\mathbf{\widetilde{F}}_{\mathcal{R}}$ need to carry consistent rotation and content representation. 
To this end, a rotation-aware consistency learning (RAC) objective (denoted as $\mathcal{L}_{RAC}$) is needed to constrain both representations.

To realize this objective, two MLPs $f_3(\cdot)$ and $f_4(\cdot)$ are used to project $\mathbf{F}_{\mathcal{R}}$ and $\mathbf{\widetilde{F}}_{\mathcal{R}}$ into embeddings, which we denote as $\mathbf{z}_{\mathcal{R}} \in \mathbb{R}^{d \times n}$ and $\mathbf{\widetilde{z}}_{\mathcal{R}}  \in \mathbb{R}^{d \times n}$.
Similarly, we have 
$\mathbf{z}'=[\mathbf{z}_{\mathcal{R}},  \mathbf{\widetilde{z}}_{\mathcal{R}}] \in \mathbb{R}^{d \times 2n}$, where $z'_j, z'_k \in \mathbf{z}'$, $z'_j, z'_k \in \mathbb{R}^{d \times 1}$, $j, k=1, \cdots, 2n$ and $j \neq k$.
Assume $z'_{j,+}$ and $z'_j$ construct a positive pair, and the rotation-aware consistency learning objective for RRoIs (denoted as $\mathcal{L}_{RAC}$) can be computed as
\begin{equation} 
\label{RRoICL}
\mathcal{L}_{RAC} =  -\sum_{j=1}^{2n} \sigma(\mathcal{R}_j) \cdot {\rm log} \frac{exp(z'_j \cdot z'_{j,+} / \tau)}{ \sum_{j=1, j \neq k}^{2n} exp(z'_j \cdot z'_{k} / \tau)}.
\end{equation}

While $\mathcal{L}_{HCL}$ in Eq.~\ref{HRoICL} is also applicable to predict horizontal bounding boxes on unseen target domains, the proposed $\mathcal{L}_{RAC}$ is especially devised for oriented bounding boxes and is able to constrain the orientation information of an oriented object detection.
Notably, it allows the pre- and post- hallucinated HRoIs and RRoIs that have the most similar orientation information to be aligned, while the rest are pushed as apart as possible.

\subsection{Style Consistency Learning}
\label{sec3.4}

For oriented objects, a domain generalized detector is supposed to show robust orientation prediction despite the great style variation.
During its training, it is necessary to constrain the style hallucination to be non-geometry, so that the original samples and the style-hallucinated samples could carry the same structural information. 
We propose a style consistency (SEC) learning to realize this objective.

To begin with, two fully-connected layers $g_1(\cdot)$ and $g_2(\cdot)$ are used to map the oriented feature representation of original sample $\mathbf{F}_{\mathcal{R}}$ and style-hallucinated sample $\mathbf{\widetilde{F}}_{\mathcal{R}}$ to category-wise probability distribution $\mathbf{p}$ and $\mathbf{\widetilde{p}}$, given by
\begin{equation} 
\label{catescore}
\mathbf{p} = g_1(\mathbf{F}_{\mathcal{R}}), \quad \mathbf{\widetilde{p}} = g_2(\mathbf{\widetilde{F}}_{\mathcal{R}}).
\end{equation}

Both category-level scores $\mathbf{p}$ and $\mathbf{\widetilde{p}}$ need to be restricted in a way such that the pixel-level category remains unchanged in both original and style diversified samples. 
For conventional restrictions such as $l$-2 norm and Kullback-Leibler (KL) divergence, the detector training can become unstable when the numerical value between $\mathbf{p}$ and $\mathbf{\widetilde{p}}$ varies greatly after hallucination.
To address this issue, the Jensen-Shannon Divergence (JSD) \cite{menendez1997jensen} is utilized to realize this style consistency objective, given by
\begin{equation} 
\label{JSDloss}
\mathcal{L}_{SEC} = \frac{1}{2}({\rm KL}[\mathbf{p}||\mathbf{\overline{p}}]+{\rm KL}[\mathbf{\widetilde{p}}||\mathbf{\overline{p}}]),
\end{equation}
where $\mathbf{\overline{p}}=(\mathbf{\widetilde{p}}+\mathbf{p})/2$, and ${\rm KL}$ denotes the Kullback-Leibler divergence.

To summarize, the proposed SEC poses a category-wise constraint between the original and style-hallucinated images, which allows the style-hallucinated images to perceive the same content of an aerial image.

\subsection{Implementation Details}

Assume a convention oriented object detector is optimized by the classification head $\mathcal{L}_{cls}$ and bounding box regression head $\mathcal{L}_{reg}$.
Then, 
allowing for Eq.~\ref{HRoICL},~\ref{RRoICL}~and~\ref{JSDloss},
the learning objective of GOOD is computed as
\begin{equation} 
\label{loss}
\mathcal{L} =  \mathcal{L}_{cls} +  \mathcal{L}_{reg}  + \mathcal{L}_{HCL} + \mathcal{L}_{RAC} + \mathcal{L}_{SEC}.
\end{equation}
For classification $\mathcal{L}_{cls}$ and regression $\mathcal{L}_{reg}$, both original feature representation and style-hallucinated representation are involved for training. 

Following the standard configuration of object detectors, ResNet-50 with Feature Pyramid Network (FPN) is used as the backbone.
In FPN, each pyramid assigns 15 anchors per location.
For the region proposal network, 1024 Region of Interests (RoIs) with a 1:3 positive to negative ratio are sampled for training. 
All the training schedules follow the default settings in \texttt{mmdetection} \footnote{https://github.com/open-mmlab/mmdetection}.

The SGD optimizer is used for training with an
initial learning rate of 0.01. The value of momentum and weight decay
is set to be 0.9 and 0.0001, respectively. 
After each decay step, the learning rate is divided by 10.
All the models are trained by a total of 20 epochs on a workstation.
4 V100 GPUs with a total batch size of 8 are used for training.

When inference, the CLIP-driven style hallucination is removed, and the detector directly predicts oriented bounding boxes on unseen target domains. 
2,000 RoIs from each pyramid level are adapted before non-maximum suppression (NMS).
A total of 2,000 RoIs are selected after NMS.

\begin{table*}[!t]
    \centering
        \footnotesize
        \caption{Experimental settings for domain generalized oriented object detection. Following prior generalized driving-scene detection works \cite{wu2022single,lin2021domain,vidit2023clip}, the shared object categories among different datasets are used to evaluate the domain generalized oriented object detector.
        }
	\resizebox{\linewidth}{!}{
	\begin{tabular}{c|c|c|c|c}
		\hline
        Setting & Source Domain & Unseen Target Domain & \#Cate. & Categories\\
        \hline
       \uppercase\expandafter{\romannumeral1} & FAIR1M  & DOTA-V1, DOTA-V1.5, DOTA-V2.0 & 10 & \makecell[c]{airplane, ship, baseball field, \\ tennis court, basketball court, football field, \\ large vehicle, small vehicle, bridge, roundabout} \\
       \hline
        \uppercase\expandafter{\romannumeral2} & SODA & DOTA-V1, DOTA-V1.5, DOTA-V2.0 & 7 & \makecell[c]{airplane, ship, storage tank, large vehicle, \\ small vehicle, helicopter, swimming pool} \\
        \hline
        \uppercase\expandafter{\romannumeral3} & DOTA-V1 & SODA, FAIR1M & 4 & \makecell[c]{airplane, ship, large vehicle, small vehicle} \\
        \hline
        \uppercase\expandafter{\romannumeral4} & DOTA-V1 & SODA, FAIR1M, HRSC & 1 & \makecell[c]{ship} \\
        \hline
	\end{tabular}
        }
	\label{experimentsetting}
\end{table*}

\begin{table*}[!t]
    \centering
        \footnotesize
         \caption{Result comparison between the proposed GOOD and existing oriented object detectors on unseen target domains and cross-domain object detectors.
        Evaluation metric $mAP_{50}$ (in \%).
        For fair evaluation, all the compared methods use the ResNet-50 backbone pre-trained on ImageNet.
        Top three results are highlighted as \colorbox{colorFst}{\bf best}, \colorbox{colorSnd}{second} and \colorbox{colorTrd}{third}, respectively.
        }
	\resizebox{\linewidth}{!}{
	\begin{tabular}{l|ccc|ccc|cc}
		\hline
		\multirow{2}{*}{Method} & \multicolumn{3}{c|}{Source Domain: FAIR1M} & \multicolumn{3}{c|}{Source Domain: SODA} & \multicolumn{2}{c}{Source Domain: DOTA} \\ 
        \cline{2-9}
        ~ & DOTA-1.0 & DOTA-1.5 & DOTA-2.0 & DOTA-1.0 & DOTA-1.5 & DOTA-2.0 & SODA & FAIR1M \\
        \hline
        \textit{One-stage:} & \\
        RetinaNet OBB \cite{lin2017focal} & 28.78 & 28.31 & 26.34 & 37.82 & 37.72 & 35.08 & 39.39 & 29.65 \\
        FCOS OBB \cite{tian2020fcos} & 30.47 & 31.46 & 29.76 & 37.49 & 39.80 & 37.45 & 42.46 & 31.86 \\
        ATSS OBB \cite{zhang2020bridging} & 35.89 & 31.12 & 29.39 & 37.38 & 38.30 & 36.41 & 41.12 & 34.67 \\
        R3Det \cite{yang2021r3det} & 35.10 & 34.79 & 33.79 & 38.77 & 39.33 & 37.39 & 44.11 & 36.51 \\
       SA2Net \cite{han2021align} & 32.31 & 32.57 & 31.41 & 35.21 & 35.70 & 33.37 & 33.25 & 29.90 \\
       GWD \cite{yang2021rethinking} & 38.59 & \rd 39.56 & 37.35 & 38.70 & 39.77 & 38.62 & 46.46 & 40.31 \\
       KLD \cite{yang2021learning} & 37.59 & 36.76 & 34.92 & 39.19 & 41.19 & 39.14 & 48.57 & 36.99 \\
       SASM \cite{hou2022shape} & 35.69 & 35.83 & 33.82 & 41.20 & 42.28 & 40.15 & 47.34 & 38.15 \\
       RepPoint \cite{li2022oriented} & 37.80 & 36.86 & 34.85 & 37.41 & 40.53 & 37.66 & 50.49 & 40.63 \\
       PSCD \cite{yu2023phase} & 33.59 & 32.41 & 30.86 & 40.42 & 39.65 & 36.70 & 43.38 & 34.72 \\
       DCFL \cite{xu2023dynamic} & \rd 39.85 & 39.07 & \rd 38.15 & 41.69 & 40.56 & 38.72 & 47.88 & 39.63 \\
       ACM \cite{xu2024rethinking} & 36.17 & 36.58 & 33.74 & 38.06 & 38.47 & 35.89 & 45.09 & 36.84 \\
        \hline
        \textit{Two-stage:} & \\
        FR OBB \cite{ren2015faster} & 33.15 & 33.17 & 30.79 & 34.98 & 36.89 & 34.89 & 44.14 & 34.46 \\
        MR OBB \cite{he2017mask} & 34.46 & 35.99 & 33.30 & 39.81 & 42.60 & \rd 41.08 & 46.32 & 36.54 \\
        Oriented RCNN \cite{xie2021oriented} & 35.72 & 34.97 & 34.19 & 40.07 & 41.23 & 38.35 & 48.83 & 37.34 \\
       RoI-Tran \cite{ding2019learning} & 36.82 & 37.02 & 35.21 & 41.93 & 43.19 & 40.92 & 52.06 & 41.18 \\
       ReDet \cite{han2021redet} & \nd 40.46 & \nd 41.06 & \nd 38.35 & \rd 42.18 & \rd 44.25 & 41.05 & \nd 54.85 & \nd 43.09 \\
       ARC \cite{pu2023adaptive} & 34.86 & 35.07 & 33.42 & 36.51 & 37.96 & 35.74 & 45.25 & 34.78 \\
       JDet \cite{xiao2024theoretically} & 37.05 & 37.28 & 35.87 & 38.46 & 39.07 & 37.21 & 47.63 & 38.97 \\
       \hline
        \textit{Cross-Domain Detector:} & \\
        S-DGOD \cite{wu2022single} & 29.37 & 30.02 & 27.86 & 36.80 & 36.51 & 34.96 & 38.05 & 30.67 \\
        DomainGen \cite{vidit2023clip} & 35.49 & 35.08 & 34.12 & 42.05 & 41.86 & 40.57 & 43.41 & 37.26 \\
        SFOD-RS \cite{liu2024clip} & 34.56 & 34.19 & 33.74 & 40.47 & 39.63 & 38.85 & 41.73 & 33.40 \\
        PDGA \cite{liu2024source} & 36.05 & 37.13 & 35.22 & 41.97 & 42.58 & 39.24 & 40.52 & 32.85 \\
        \hline
        \textbf{GOOD} (RoI-Tran) & 37.94 & 37.55 & 36.31 & \nd 44.03 & \nd 45.86 & \nd 42.14 & \rd 54.11 & \rd 42.74 \\
        ~ & \textcolor{red}{$\uparrow$1.12} &  \textcolor{red}{$\uparrow$0.53} & \textcolor{red}{$\uparrow$1.10} & \textcolor{red}{$\uparrow$2.10} & \textcolor{red}{$\uparrow$2.67} & \textcolor{red}{$\uparrow$1.22} & \textcolor{red}{$\uparrow$2.05} & \textcolor{red}{$\uparrow$1.56}\\
         \textbf{GOOD} (ReDet) & \fs 42.61 & \fs 42.70 & \fs 40.36 & \fs 44.65 & \fs 47.54 & \fs 43.88 & \fs 57.19 & \fs 46.21 \\
         ~ & \textcolor{red}{$\uparrow$2.15} & \textcolor{red}{$\uparrow$1.64} & \textcolor{red}{$\uparrow$2.01} & \textcolor{red}{$\uparrow$2.47} & \textcolor{red}{$\uparrow$3.29} & \textcolor{red}{$\uparrow$2.83} & \textcolor{red}{$\uparrow$3.08} & \textcolor{red}{$\uparrow$3.12} \\
        \hline
	\end{tabular}
        }
	\label{mergeallresult}
\end{table*}

\section{Experiment}
\label{sec4}

\subsection{Dataset \& Evaluation Protocols}

\subsubsection{Datasets}

\noindent \textbf{FAIR1M} \cite{sun2022fair1m} is a fine-grained aerial object detection dataset with 37 corresponding fine-grained categories.
It has more than 15,000 images and 1 million instances.
To set up a domain generalization setting which exists same object categories from cross-domain, we merge the 11 fine-grained airplane categories into a single \textit{airplane} category, 9 fine-grained ship categories into a \textit{ship} category, 10 fine-grained vehicle categories into \textit{large vehicle} and \textit{small vehicle} categories, respectively.
On the other hand, as the test set of FAIR1M is not publicly available, we randomly split the official training set into an 8:2 ratio, designating the source domain for training and the target domain for inference, respectively.
The images are split into 1024$\times$1024 for training, and are merged for evaluation. The overlap between two cropped image patches is 200$\times$200.

\noindent \textbf{DOTA} \cite{xia2018dota} consists of 2,806 aerial images with 188,282 instances. The annotation of DOTA \footnote{https://captain-whu.github.io/DOTA/index.html} has three versions, namely, DOTA-V1.0, DOTA-V1.5 and DOTA-V2.0 \cite{ding2021object}.
DOTA-V1.0 has 15 object categories. DOTA-V1.5 and DOTA-V2.0 has one and three more object categories than DOTA-V1.0, respectively.
Following the default setting of MMRotate \footnote{https://github.com/open-mmlab/mmrotate}, the images are split into 1024$\times$1024 for training with an overlap of 200$\times$200, and are merged for evaluation. 

\noindent \textbf{SODA} \cite{Cheng2023SODA} consists of 2,513 images with up to 872,069 instances from 9 classes. 
It is an oriented object detection benchmark for small objects. 
All the data split and pre-processing follow the default setting of MMRotate \footnote{https://github.com/shaunyuan22/SODA-mmrotate}. The patch resolution is 1024$\times$1024, and the overlap is 200$\times$200.

\noindent \textbf{HRSC} \cite{liu2016ship} is a ship detection dataset with arbitrary orientations.
The training and testing split directly follows the default setting of MMRotate, where all the images are resized to 1024$\times$1024 for both training and testing.

\subsubsection{Evaluation Protocols}

The object category definition and number between SODA \cite{Cheng2023SODA}, FAIR1M \cite{sun2022fair1m}, DOTA \cite{xia2018dota} and HRSC \cite{liu2016ship} is different.
Some categories in FAIR1M \cite{sun2022fair1m} do not occur in the rest three datasets, and vice versa.
Following prior domain generalized driving-scene object detection \cite{wu2022single,lin2021domain,vidit2023clip}, the common object categories among different datasets are used to evaluate the domain generalized object detector.
The four generalization settings and the corresponding object categories are listed in Table~\ref{experimentsetting}.

Specifically, the object category number and the object categorization system is different among these four domains.
The common object category number is 10, 7, 4 and 1, when we involve two, two, three and four of the oriented object detection datasets, namely, DOTA, SODA, FAIR1M and HRSC. 
To highlight, if all four domains are involved, there is only one common category, which is too specialized for ship detection. If three domains are involved, there are only four common categories.
They are still much less than the in-domain object detection tasks, where there are usually tens of object categories. Therefore, a trade-off between domain diversity and object category number has to be made.
The first and second settings do a trade-off for more common object categories from three datasets, while the third and fourth settings do a trade-off for more domain diversity, including all four datasets.

For all the experiments, Mean Average Precision (mAP) is used as the evaluation metric. 
In line with \cite{xia2018dota}, mAP@0.5 is reported, which classifies a prediction as a true positive when it aligns with the ground-truth label and achieves an Intersection over Union (IOU) score of over 0.5 with the ground-truth bounding box.

\subsection{Comparison with State-of-the-art}

\begin{table*}[!t]
\begin{center}
\caption{Per-category performance comparison between the proposed GOOD and the state-of-the-art methods. 
FAIR1M dataset is used as the source domain.
DOTA-V1.0 is used as the unseen target domain.
Evaluation metric $mAP_{50}$ (in percentage \%).
AP: airplane; BF: baseball-field; BG: Bridge; SV: small-vehicle; LV: large-vehicle; SP: ship; TC: tennis-court; BC: basketball-court; FF: football-field; RA: roundabout.
Top three results are highlighted as \colorbox{colorFst}{\bf best}, \colorbox{colorSnd}{second} and \colorbox{colorTrd}{third}, respectively.
}
\resizebox{\linewidth}{!}{
\begin{tabular}{l|cccccccccc|c}
\hline
Method & AP & BF & BG & SV & LV & SP & TC & BC & FF & RA & mAP\\ 
\hline
R3Det \cite{yang2021r3det} & 59.36 & 34.08 & 10.25 & \nd 16.63 & 3.95 & 65.50 & 68.69 & 22.43 & 42.53 & \fs 27.76 & 35.10 \\
RepPoint \cite{li2022oriented} & 64.32 & 38.12 & 12.76 & \fs 21.32 & 5.12 & 69.93 & \rd 76.45 & 31.59 & 39.01 & 19.27 & 37.80 \\
RoI-Tran \cite{ding2019learning} & 60.64 & \nd 41.04 & \nd 13.63 & 12.34 & \rd 5.58 & \rd 76.02 & 62.32 & 29.32 & \nd 45.32 & 22.23 & 36.82 \\
PSCD \cite{yu2023phase} & 56.78 & 37.92 & 12.63 & 9.75 & 3.86 & 72.53	& 58.96	& 27.04	& 42.16	& 18.14
 & 33.59 \\
DCFL \cite{xu2023dynamic} & \rd 75.85 & 39.71 & \rd 13.19 & 12.86 & 4.57 & 75.82 & 71.52 & \rd 34.86 & \rd 44.95 & 23.61 & \rd 39.85 \\
ACM \cite{xu2024rethinking} & 59.31 & \rd 40.87 & 13.08 & 11.69 & 5.28 & 74.68 & 61.75 & 30.17 & 44.63 & 20.39
 & 36.17 \\
ARC \cite{pu2023adaptive} & 58.94 & 38.96 & 12.05 & 12.34 & 4.25 & 73.39 & 58.89 & 27.86 & 43.04 & 19.67 & 34.86 \\
JDet \cite{pu2023adaptive} & 67.93 & 40.15 & 11.87 & 13.07 & 4.86 & 74.21 & 63.06 & 29.78 & 42.68 & 24.05 & 37.05 \\
S-DGOD \cite{wu2022single} & 50.62 & 33.58 & 10.12 & 8.69 & 2.52 & 63.37 & 54.18 & 22.54 & 36.59 & 13.64 & 29.37 \\
DomainGen \cite{vidit2023clip} & 59.37 & 40.61 & 12.58 & 11.45 & 5.04 & 74.56 & 60.92 & 28.02 & 43.96 & 20.58 & 35.49 \\
SFOD-RS \cite{liu2024clip} & 57.89 & 38.58 & 11.05 & 10.68 & 4.39 & 74.69 & 60.17 & 27.81 & 42.87 & 20.76 & 34.56 \\
PDGA \cite{liu2024source} & 60.95 & 40.76 & 12.94 & 12.72 & 5.36 & 74.84 & 61.62 & 28.03 & 43.45 & 21.02 & 36.05 \\
ReDet \cite{han2021redet} & \nd 77.89 & 34.63 & 12.52 & 10.58 & \fs 7.62 & \fs 76.77 & \fs 77.05 & \nd 41.68 & 40.36 & \rd 25.18 & \nd 40.46 \\
\hline
\textbf{GOOD} (Ours) & \fs 78.23 & \fs 41.35 & \fs 16.34 & \rd 13.32 & \nd 5.93 & \nd 76.63 & \nd 76.54 & \fs 45.64 & \fs 47.12 & \nd 25.43 & \fs 42.61 \\
\hline
\end{tabular}
}
\label{percate}
\end{center} 
\end{table*}

\begin{table*}[!t]
\begin{center}
\caption{Per-category performance comparison between the proposed GOOD and the state-of-the-art methods. 
FAIR1M dataset is used as the source domain.
DOTA-V1.5 is used as the unseen target domain.
Evaluation metric $mAP_{50}$ (in percentage \%).
AP: airplane; BF: baseball-field; BG: Bridge; SV: small-vehicle; LV: large-vehicle; SP: ship; TC: tennis-court; BC: basketball-court; FF: football-field; RA: roundabout.
Top three results are highlighted as \colorbox{colorFst}{\bf best}, \colorbox{colorSnd}{second} and \colorbox{colorTrd}{third}, respectively.
}
\resizebox{\linewidth}{!}{
\begin{tabular}{l|cccccccccc|c}
\hline
Method & AP & BF & BG & SV & LV & SP & TC & BC & FF & RA & mAP\\ 
\hline
R3Det \cite{yang2021r3det} & 59.34 & 34.13 & 10.14 & 27.04 & 3.61 & 64.87 & 68.85 & 23.56 & 41.58 & \fs 14.78 & 34.79 \\
RepPoint \cite{li2022oriented} & 63.85 & 37.76 & 12.92 & \fs 32.75 & 4.79 & 67.54 & \rd 74.76 & 30.07 & 40.24 & 3.87 & 36.86 \\
RoI-Tran \cite{ding2019learning} & 60.62 & \nd 41.07 & 13.43 & \rd 29.82 & \rd 5.82 & 70.31 & 62.06 & 29.37 & \rd 44.36 & \rd 12.89 & 37.02 \\
PSCD \cite{yu2023phase} & 56.43 & 35.86 & 10.26 & 25.05 & 4.76 & 63.79 & 57.28 & 25.68 & 38.69 & 9.26 & 32.41 \\
DCFL \cite{xu2023dynamic} & \rd 72.86 & 40.14 & \rd 14.37 & 28.69 & 5.28 & \rd 74.18 & 66.27 & \rd 31.59 & \nd 44.57 & 12.68 & \rd 39.07 \\
ACM \cite{xu2024rethinking} & 66.04 & \rd 40.25 & 12.86 & 27.47 & 4.65 & 71.23 & 63.58 & 25.86 & 42.04 & 11.54 & 36.58 \\
ARC \cite{pu2023adaptive} & 64.13 & 36.42 & 12.51 & 25.32 & 4.08 & 65.06 & 61.96 & 28.17 & 41.65 & 11.71 & 35.07 \\
JDet \cite{xiao2024theoretically} & 71.76 & 38.67 & 13.45 & 27.03 & 5.03 & 71.34 & 65.12 & 29.05 & 39.59 & 10.67 & 37.28 \\
S-DGOD \cite{wu2022single} & 53.35 & 34.26 & 10.29 & 20.58 & 4.36 & 62.58 & 54.59 & 23.08 & 32.43 & 11.35 & 30.02 \\
DomainGen \cite{vidit2023clip} & 63.88 & 37.58 & 12.16 & 26.36 & 5.15 & 69.74 & 58.73 & 29.47 & 35.62 & 12.26 & 35.08 \\
SFOD-RS \cite{liu2024clip} & 61.29 & 35.84 & 11.48 & 25.51 & 4.84 & 68.26 & 60.47 & 27.96 & 34.15 & 10.78 & 34.19 \\
PDGA \cite{liu2024source} & 72.63 & 38.82 & \nd 14.53 & 26.85 & 5.07 & 70.67 & 64.78 & 28.61 & 40.78 & 11.26 & 37.13 \\
ReDet \cite{han2021redet} & \nd 77.95 & 37.82 & 12.18 & 29.34 & \fs 7.03 & \fs 77.05 & \fs 77.28 & \nd 39.48 & 41.08 & 11.29 & \nd 41.06 \\
\hline
\textbf{GOOD} (Ours) & \fs 78.04 & \fs 41.54 & \fs 15.93 & \nd 30.58 & \nd 6.14 & \nd 76.82 & \nd 76.43 & \fs 42.33 & \fs 46.42 & \nd 13.13 & \fs 42.70 \\
\hline
\end{tabular}
}
\label{percate15}
\end{center} 
\end{table*}

\begin{table*}[!t]
\begin{center}
\caption{Per-category performance comparison between the proposed GOOD and the state-of-the-art methods. 
FAIR1M dataset is used as the source domain.
DOTA-V2.0 is used as the unseen target domain.
Evaluation metric $mAP_{50}$ (in percentage \%).
AP: airplane; BF: baseball-field; BG: Bridge; SV: small-vehicle; LV: large-vehicle; SP: ship; TC: tennis-court; BC: basketball-court; FF: football-field; RA: roundabout.
Top three results are highlighted as \colorbox{colorFst}{\bf best}, \colorbox{colorSnd}{second} and \colorbox{colorTrd}{third}, respectively.
}
\resizebox{\linewidth}{!}{
\begin{tabular}{l|cccccccccc|c}
\hline
Method & AP & BF & BG & SV & LV & SP & TC & BC & FF & RA & mAP\\ 
\hline
R3Det \cite{yang2021r3det} & 57.85 & 32.03 & 10.02 & 26.21 & 3.45 & 61.78 & 68.53 & 23.41 & 41.02 & \fs 13.58 & 33.79 \\
RepPoint \cite{li2022oriented} & 60.24 & 32.09 & 12.21 & \nd 28.55 & 4.62 & 63.67 & 73.42 & 29.19 & 38.94 & 3.52 & 34.85 \\
RoI-Tran \cite{ding2019learning} & 60.03 & \nd 33.76 & 12.60 & 23.63 & 5.01 & 69.46 & 61.76 & 29.11 & \nd 44.02 & \nd 12.69 & 35.21 \\
PSCD \cite{yu2023phase} & 54.82 & 21.69 & 8.45 & 20.58 & 4.39 & 60.71 & 62.35 & 29.68 & 36.16 & 5.37 & 30.86 \\
DCFL \cite{xu2023dynamic} & \rd 68.16 & 31.54 & \nd 14.08 & 27.79 & 4.02 & \rd 73.62 & \rd 74.37 & \rd 37.65 & \rd 42.59 & 6.98 & \rd 38.15 \\
ACM \cite{xu2024rethinking} & 59.73 & 28.76 & 12.09 & 25.86 & 4.75 & 65.89 & 68.92 & 29.19 & 37.58 & 4.28 & 33.74 \\
ARC \cite{pu2023adaptive} & 60.26 & 29.98 & 11.47 & 24.51 & 5.23 & 63.74 & 66.86 & 28.83 & 36.74 & 5.27 & 33.42 \\
JDet \cite{xiao2024theoretically} & 64.34 & \rd 32.82 & \rd 13.19 & 21.87 & \rd 5.54 & 69.58 & 69.67 & 31.02 & 41.37 & 6.10 & 35.87 \\
S-DGOD \cite{wu2022single} & 52.65 & 19.54 & 7.87 & 20.73 & 4.18 & 61.53 & 58.96 & 27.89 & 32.63 & 5.09 & 27.86 \\
DomainGen \cite{vidit2023clip} & 63.90 & 31.76 & 12.26 & 22.95 & 4.58 & 66.39 & 68.21 & 30.74 & 39.82 & 6.54 & 34.12 \\
SFOD-RS \cite{liu2024clip} & 62.67 & 29.85 & 11.72 & 21.27 & 4.35 & 65.37 & 63.49 & 31.58 & 40.95 & 5.67 & 33.74 \\
PDGA \cite{liu2024source} & 67.57 & 30.51 & 12.93 & 26.84 & 4.97 & 70.16 & 66.28 & 30.13 & 39.69 & 5.84 & 35.22 \\
ReDet \cite{han2021redet} & \fs 75.15 & 32.45 & 11.45 & \rd 28.09 & \fs 6.83 & \nd 75.14 & \fs 76.73 & \nd 38.92 & 40.34 & 8.32 & \nd 38.35 \\
\hline
\textbf{GOOD} (Ours) & \nd 70.44 & \fs 33.84 & \fs 15.63 & \fs 29.62 & \nd 6.04 & \fs 75.25 & \nd 76.04 & \fs 41.93 & \fs 45.81 & \rd 9.43 & \fs 40.36 \\
\hline
\end{tabular}
}
\label{percate2}
\end{center} 
\end{table*}

\begin{table}[!t]
    \centering
        \footnotesize
        \caption{Result comparison between the proposed GOOD and baseline on the setting: DOTA-V1.0$\rightarrow$\{SODA, FAIR1M, HRSC\}. Evaluation metric $mAP_{50}$ presented in percentage. 
        }
	\begin{tabular}{c|ccc}
		\hline
		\multirow{2}{*}{Method} & \multicolumn{3}{c}{Target Domains} \\ 
        \cline{2-4}
        ~ & SODA & FAIR1M & HRSC \\
        \hline
       RoI-Tran \cite{ding2019learning} & 67.20 & 35.05 & 35.94 \\
       \textbf{GOOD} (RoI-Tran) & \fs 69.92 & \fs 39.26 & \fs 38.87 \\
       ~ & \textcolor{red}{$\uparrow$2.72} &  \textcolor{red}{$\uparrow$4.21} &  \textcolor{red}{$\uparrow$2.93} \\
       \hline
       ReDet \cite{han2021redet} & 75.89 & 46.37 & 59.71 \\
       \textbf{GOOD} (ReDet) & \fs 78.97 & \fs 48.44 & \fs 62.08 \\
        ~ & \textcolor{red}{$\uparrow$3.08} &  \textcolor{red}{$\uparrow$2.07} &  \textcolor{red}{$\uparrow$2.37} \\
        \hline
	\end{tabular}
	\label{HRSCsource}
\end{table}

The proposed GOOD along with extensive state-of-the-art oriented object detectors are involved for experimental evaluation.
Specifically, twelve one-stage oriented object detection methods (namely, RetinaNet OBB \cite{lin2017focal}, FCOS OBB \cite{tian2020fcos}, ATSS OBB \cite{zhang2020bridging}, R3Det \cite{yang2021r3det}, SA2Net \cite{han2021align}, GWD \cite{yang2021rethinking}, KLD \cite{yang2021learning}, SASM \cite{hou2022shape}, RepPoint \cite{li2022oriented}, PSCD \cite{yu2023phase}, DCFL \cite{xu2023dynamic}, ACM \cite{xu2024rethinking}) and seven two-stage oriented object detection methods (namely, FR OBB \cite{ren2015faster}, MR OBB \cite{he2017mask}, Oriented RCNN \cite{xie2021oriented}, RoI-Tran \cite{ding2019learning}, ReDet \cite{han2021redet},  ARC \cite{pu2023adaptive}, JDet \cite{xiao2024theoretically}) are included.
For boarder comparison, four recent domain generalized object detectors are also involved for comparison, namely, S-DGOD \cite{wu2022single}, DomainGen \cite{vidit2023clip}, SFOD-RS \cite{liu2024clip} and PDGA \cite{liu2024source}. The oriented detection head is modified on S-DGOD and DomainGen, as they are not initially devised for oriented object detection.

For fair comparison, all the detectors are trained using the ResNet-50 backbone with a pre-trained weight on ImageNet. 
Besides, under each domain generalization setting, each detector is implemented under the corresponding default parameter settings with the official source code.

\begin{figure*}[!t]
    \centering 
    \subfigure[RoI-Trans]{
    \includegraphics[width=1.5in]{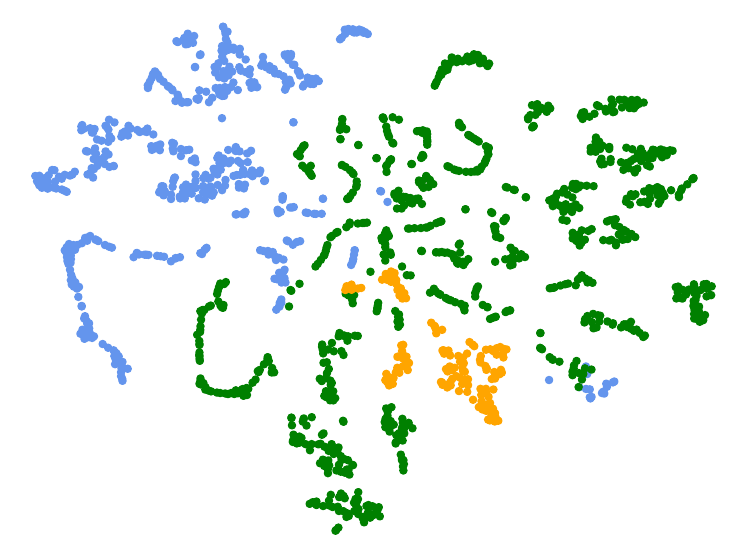}
    }
    \hfill
    \subfigure[GOOD (RoI-Trans)]{
    \includegraphics[width=1.5in]{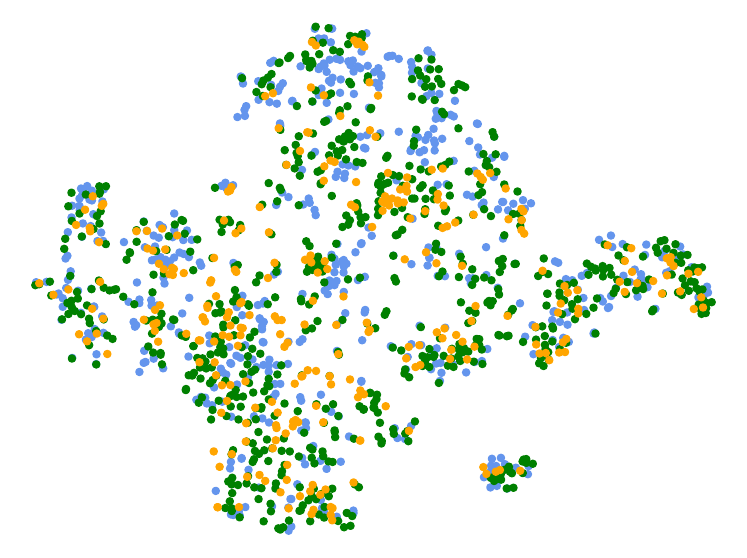}
    } 
    \hfill
    \subfigure[ReDet]{
    \includegraphics[width=1.5in]{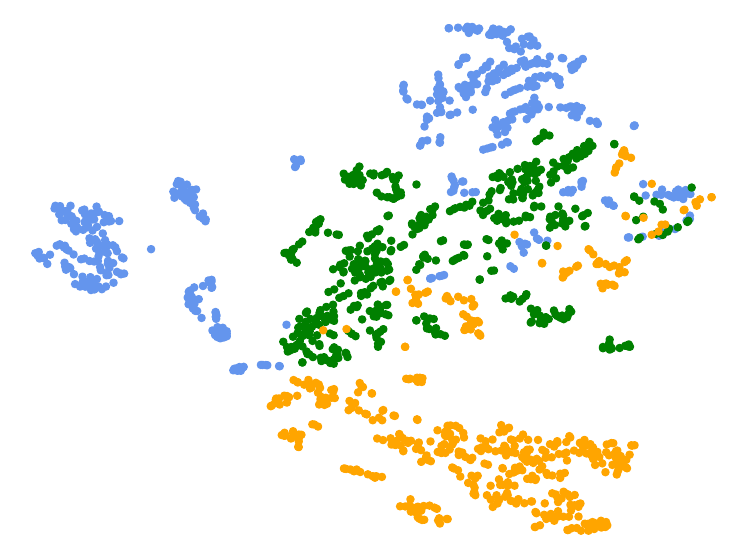}
    }
    \hfill
    \subfigure[GOOD (ReDet)]{
    \includegraphics[width=1.5in]{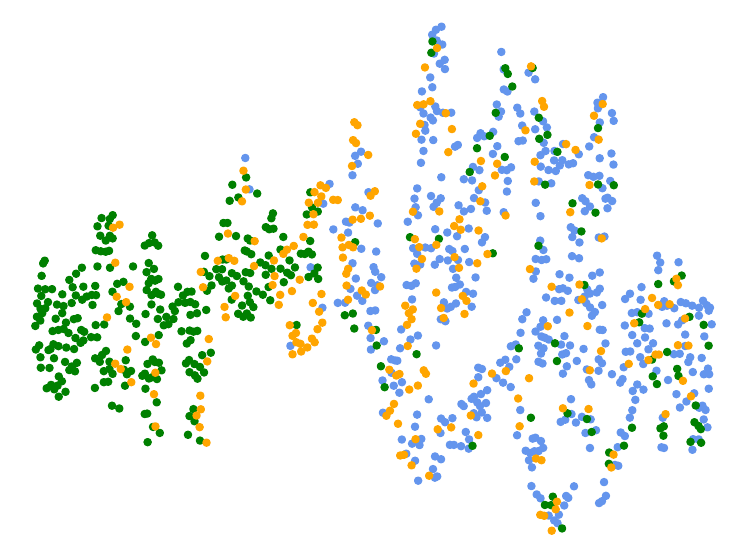}
    } 
   \caption{t-SNE visualization of the cross-domain feature space. Blue, green and orange denotes the target domains of SODA, FAIR1M, HRSC, respectively. The more uniformly distributed from these domains, the better feature generalization.}  
   \label{tsne}   
\end{figure*}

\subsubsection{FAIR1M as Source Domain}

The second column of Table~\ref{mergeallresult} reports the results.
The proposed GOOD yields an mAP of 42.61\%, 42.70\% and 40.36\% on DOTA-1.0, -1.5 and -2.0, respectively, 
outperforming all the compared oriented object detectors. 
When embedded into ReDet \cite{han2021redet}, it outperforms the ReDet baseline, which is the second best on these three unseen target domains, by 2.15\%, 1.64\% and 2.01\% mAP, respectively.
Besides, it significantly outperforms the third-best GWD \cite{yang2021rethinking}, which only yields an mAP of 38.59\%, 39.56\% and 37.35\% on DOTA-1.0, -1.5 and -2.0, respectively.
When embedded into RoI-Tran \cite{ding2019learning}, it outperforms the RoI-Tran baseline by 1.12\%, 0.53\% and 1.10\% mAP, respectively. 
To better understand how GOOD improves the oriented object detection performance on unseen target domains, we provide a break-down analysis on each object category.

\paragraph{FAIR1M $\rightarrow$ DOTA-1.0}
Table~\ref{percate}~reports the results.
The proposed GOOD outperforms the second-best by 2.15\% mAP.
In terms of the per-category average precision, the proposed GOOD outperforms the original ReDet on seven out of ten categories, only showing slightly inferior performance on LV, SP and TC.
Finally, the proposed GOOD shows the best performance on five out of ten categories.

\paragraph{FAIR1M $\rightarrow$ DOTA-1.5}
Table~\ref{percate15}~reports the results.
The proposed GOOD outperforms the second-best by 1.64\% mAP.
Besides, compared with the original ReDet, the proposed GOOD outperforms it on seven out of ten categories.
Finally, the proposed GOOD shows the best performance on five out of ten categories when compared with all the runner-ups.
Notably, the category-wise average precision improvement on BF, BG, BC and FF is 3.72\%, 3.75\%, 2.85\% and 5.34\%, respectively.

\paragraph{FAIR1M $\rightarrow$ DOTA-2.0}
Table~\ref{percate2}~reports the results.
The proposed GOOD outperforms the second-best by up to 2.01\% mAP.
Compared with the original ReDet, the proposed GOOD is able to outperform seven out of ten categories.
Further, it achieves the best performance on six out of ten categories when compared with all the runner-ups.

In summary, the proposed GOOD allows the baseline to generalize better on seven out of the ten categories, by up to 6.7\% mAP on FF category, 4.8\% mAP on BF category, respectively. 
On the other hand, a worse performance than baseline is demonstrated on three categories (LV, SP, TC) on all the three unseen target domains.
It may be explained that, the density, shape and size statistic of LV, SP, TC \cite{xia2018dota,ding2021object} is more varied than other categories. The proposed GOOD only constrains the orientation and content but may over-fit the shape, making it less qualified to predict the precise boxes on objects with varied unseen shapes.

\subsubsection{SODA as Source Domain}

The third column of Table~\ref{mergeallresult} reports the results.
The proposed GOOD yields an mAP of 44.65\%, 47.54\% and 43.88\% on DOTA-1.0, -1.5 and -2.0, 
outperforming all the compared oriented object detectors.
Specifically, when embedded into ReDet, the proposed GOOD outperforms the ReDet baseline by 2.47\%, 3.29\% and 2.83\% mAP on these three unseen target domains, respectively. 
When embedded into RoI-Tran, the proposed GOOD achieves the second-best, outperforming the baseline by 2.10\%, 2.67\% and 1.22\% mAP on these three unseen target domains, respectively. 

\subsubsection{DOTA-V1 as Source Domain}

The forth column of Table~\ref{mergeallresult} reports the results.
The proposed GOOD achieves the best generalization results, yielding an mAP of 57.19\% and 46.21\% on SODA and FAIR1M, respectively.
When embedded into ReDet, the proposed GOOD is able to outperform the baseline by 3.08\% and 3.12\$ mAP on both unseen target domains.
When embedded into RoI-Tran, the proposed GOOD also outperforms the baseline by 2.05\% and 1.56\% mAP on both unseen target domains. 

\subsubsection{Generalized Ship Detection}

Table~\ref{HRSCsource}~reports the results when detecting only the ship category on unseen target domains. 
The proposed GOOD achieves the best results on all three unseen target domains, yielding an mAP of 78.97\%, 48.44\% and 62.08\%, respectively. 
Specifically, when embedded into ReDet, the proposed GOOD outperforms the baseline by 3.08\%, 2.07\% and 2.37\% mAP on the SODA, FAIR1M and HRSC unseen target domains, respectively.
Similarly, when embedded into RoI-Tran, the proposed GOOD outperforms the baseline by 2.72\%, 4.21\% and 2.93\% mAP on the SODA, FAIR1M and HRSC unseen target domains, respectively.

\subsection{Comparison with Other Related Techniques}

\subsubsection{Pre-training Schemes}

The proposed GOOD uses the frozen CLIP image encoder to hallucinate the styles. We compare it with other pre-training image encoders from ImageNet \cite{deng2009imagenet}, MoCo \cite{chen2020improved}, MAE \cite{he2022masked} and BYOL \cite{grill2020bootstrap}. The experiments are conducted under the SODA$\rightarrow$\{DOTA-V1.0, -V1.5, V2.0\} setting. ReDet \cite{han2021redet} is used as the baseline model. 
Besides, replacing the ImageNet pre-trained encoder in ReDet with CLIP encoder is also compared to inspect if there is data leakage.
For fair evaluation, all the pre-trained image encoders are ResNet-50.
Table~\ref{encoderresult}~reports the outcomes. The proposed GOOD outperforms these methods by at least 1\% mAP on all these unseen target domains.

\begin{table}[!t]
    \centering
        \footnotesize
        \caption{Result comparison between the frozen CLIP image encoder, simple feature normalization and other pre-training schemes on the setting: SODA$\rightarrow$\{DOTA-V1.0, -V1.5, V2.0\}. Evaluation metric $mAP_{50}$ presented in percentage.
        }
	\begin{tabular}{c|ccc}
		\hline
		\multirow{2}{*}{Method} & \multicolumn{3}{c}{Target Domains} \\ 
        \cline{2-4}
         ~ & D-1.0 & D-1.5 & D-2.0 \\
        \hline
        ReDet & 42.18 & 44.25 & 41.05 \\
        ReDet w. CLIP & 42.93 & 44.31 & 41.84 \\
        \hline
        ImageNet \cite{deng2009imagenet} & 42.56 & 43.79 & 41.23 \\
        MoCo \cite{chen2020improved} & 42.83 & 44.60 & 41.54 \\
        BYOL \cite{grill2020bootstrap} & 43.24 & 45.31 & 42.17 \\
        MAE \cite{he2022masked} & 43.06 & 44.58 & 42.31 \\
        \hline
        instance norm. \cite{ulyanov2016instance} & 42.97 & 45.06 & 41.96 \\
        layer norm. \cite{ba2016layer} & 42.65 & 44.93 & 42.02 \\
        \hline
        GOOD (ReDet) & \fs 44.65 & \fs 47.54 & \fs 43.88  \\
        \hline
	\end{tabular}
	\label{encoderresult}
\end{table}

\subsubsection{Simple Augmentation Techniques}

To validate the effectiveness of the CLIP image encoder, we further compare the proposed GOOD with simpler feature normalization methods, namely, instance normalization \cite{ulyanov2016instance} and layer normalization \cite{ba2016layer}. 
They are denoted as instance norm. and layer norm., respectively.
Specifically, both methods directly implement on the source image branch, and the style hallucination has been removed. Table~\ref{encoderresult}~reports the outcomes. 
The proposed GOOD significantly outperforms both methods, indicating its effectiveness to enrich the styles for domain generalization.

\subsection{Discussion}

\subsubsection{Computational Cost Analysis}

Table~\ref{computation}~reports the training speed, testing speed and model size (in million, M) of the proposed GOOD and the ROI-Tran baseline~\cite{ding2019learning}. Following~\cite{ding2019learning}, all the experiments are tested on images with size of 1024$\times$1024 on a single TITAN X, and the performance of RoI-Trans is directly cited. 
The experiments are conducted under the SODA$\rightarrow$\{DOTA-V1.0, -V1.5, V2.0\} setting. The proposed GOOD does not increase any trainable parameter number or inference time when compared with the baseline, but the training time per image is slower.
This is because the proposed GOOD uses style hallucination to enrich the style diversity of each feature embedding, where the encoders and other modules are frozen. Therefore, it does not introduce additional parameter number, but increases the training time. 
Not introducing additional parameter number is very promising to adapt existing oriented object detectors to various unseen domains.

\begin{table}[!t]
    \centering
        \footnotesize
        \caption{Computational cost analysis between the proposed GOOD and ROI-Tran baseline. Following~\cite{ding2019learning}, all the speed are tested on images with size of 1024$\times$1024
        on a single TITAN X. \#param.: parameter number (in million).
        }
	\begin{tabular}{c|ccc}
		\hline
		Method & train speed & test speed & \#param. \\ 
        \hline
        RoI-Tran & 0.475 & 0.170 & 245M \\
        GOOD (RoI-Tran) & 0.728 & 0.170 & 245M \\ 
         \hline
	\end{tabular}
	\label{computation}
\end{table}

\subsubsection{In-Domain Performance Analysis}

It would be interesting to further study if the proposed GOOD can improve the in-domain performance of oriented object detection. The ten-category DOTA-V1.0, -V1.5 and -V2.0 (setting \uppercase\expandafter{\romannumeral1} in Table~\ref{experimentsetting}) are still used for inference, but we use the training set of DOTA-V1.0, -V1.5 and -V2.0 to train the oriented object detector. In this way, the training and inference images are within the same domain. The baseline models include both RoI-Tran \cite{ding2019learning} and ReDet \cite{han2021redet}. The proposed GOOD is embedded into both baselines for comparison. The results are shown in Table~\ref{indomain}. It can be seen that, the proposed GOOD (RoI-Trans baseline) shows an mAP improvement of 2.82\%, 2.65\% and 2.33\% on DOTA-V1.0, -V1.5 and -V2.0. Similarly, the proposed GOOD (ReDet baseline) shows an mAP improvement of 2.22\%, 2.42\% and 2.65\% on DOTA-V1.0, -V1.5 and -V2.0. These outcomes indicate the effectiveness of GOOD on the in-domain setting. 

\begin{table}[!t]
    \centering
        \footnotesize
        \caption{In-domain performance analysis between the proposed GOOD and baseline. Evaluation metric $mAP_{50}$ presented in percentage.
        }
	\begin{tabular}{c|ccc}
		\hline
		\multirow{2}{*}{Method} & \multicolumn{3}{c}{In-domain $mAP_{50}$} \\ 
        \cline{2-4}
         ~ & D-1.0 & D-1.5 & D-2.0 \\
        \hline
         RoI-Tran & 71.72 & 73.17 & 57.75 \\
         GOOD (RoI-Tran) & \fs 74.54 & \fs 75.82 & \fs 60.08 \\ 
         \hline
        ReDet & 78.28 & 79.94 & 59.32 \\
        GOOD (ReDet) & \fs 80.50 & \fs 82.36 & \fs 61.97  \\
        \hline
	\end{tabular}
	\label{indomain}
\end{table}

\subsubsection{Difference from Knowledge Distillation}

The proposed GOOD is a typical style hallucination based method. It is essentially different from the knowledge distillation paradigm in multiple predominant ways which prevent the data leakage. Firstly, the teacher-student structure requires the training of both the teacher and the student branch, as the teacher net needs to first learn the knowledge online. In contrast, the auxiliary branch in our method directly shares the modules and trainable parameters from the training branch on the source images. Secondly, the teacher-student structure requires the weight interaction to conduct knowledge transfer. In contrast, there is no such weight interaction between two branches in the proposed method. Thirdly, in the the teacher-student structure, the student net has to learn from the teacher's soft prediction for knowledge transfer. In contrast, the training branch and auxiliary branch of the proposed method are both supervised by the ground truth map.

On the other hand, style hallucination is a typical paradigm to tackle domain generalization \cite{lee2022wildnet,zhao2024style,yi2024hallucinated}. The objective of a domain generalization model is to show well generalization to the target domains without accessing it during training. The hallucinated styles are general, not specific for the images from a certain target domain. 
Therefore, they are essentially treated as a kind of augmentation, rather than leak the knowledge from the target domain. Specified in our context, the CLIP image encoder definitely contains more abundant style information, as it leverages the huge amount of images from the Internet for pre-training.
However, the Internet images are also general and are far different from the aerial images collected from the high-resolution satellite sensors. The outcomes in Table~\ref{encoderresult} support this point. Naively replacing the ImageNet pre-trained encoder by CLIP does not lead to a significant performance improvement on unseen target domains. 

\subsection{Ablation Studies}

\begin{table}[!t]
    \centering
        \footnotesize
        \caption{Ablation study on each component. Experimental setting: SODA$\rightarrow$\{DOTA-V1.0, -V1.5, V2.0\}. Evaluation metric $mAP_{50}$.
        Top three results are highlighted as \colorbox{colorFst}{\bf best}, \colorbox{colorSnd}{second} and \colorbox{colorTrd}{third}, respectively.
        }
	\begin{tabular}{cccc|ccc}
		\hline
		\multicolumn{4}{c|}{Component} & \multicolumn{3}{c}{Target Domains} \\ 
        \hline
        Style & $\mathcal{L}_{HCL}$ & $\mathcal{L}_{RCL}$ & $\mathcal{L}_{SEC}$ & D-1.0 & D-1.5 & D-2.0 \\
        \hline
       & & & & 42.18 & 44.25 & 41.05 \\
        $\checkmark$ & & & & 43.17 & 45.16 & 41.62 \\
        $\checkmark$ & $\checkmark$ & & & \rd 43.24 & \rd 45.77 & \rd 42.08 \\ 
        $\checkmark$ & $\checkmark$ & $\checkmark$ & & \nd 43.54 & \nd 46.49 & \nd 42.49 \\
        $\checkmark$ & $\checkmark$ & $\checkmark$ & $\checkmark$ & \fs 44.65 & \fs 47.54 & \fs 43.88 \\
        \hline
	\end{tabular}
	\label{ablation1}
\end{table}

\begin{table}[!t]
    \centering
        \footnotesize
        \caption{Impact on rotation angle prediction. Experimental setting: SODA$\rightarrow$\{DOTA-V1.0, -V1.5, V2.0\}. Evaluation metric $RMSD$ (root-mean-square deviation) is computed between the rotation angle prediction and ground truth of each object. Results in radian (rad). The less RMSD, the better orientation. 
        }
	\begin{tabular}{c|ccc}
		\hline
		\multirow{2}{*}{Method} & \multicolumn{3}{c}{RMSD $\downarrow$ on Target Domains} \\ 
        \cline{2-4}
         ~ & D-1.0 & D-1.5 & D-2.0 \\
        \hline
        RoI-Tran & 0.116 & 0.078 & 0.130 \\
        GOOD (RoI-Tran) & \fs 0.074 & \fs 0.045 & \fs 0.079 \\ 
         ~ & \textcolor{red}{$\downarrow$36.2\%} & \textcolor{red}{$\downarrow$42.3\%}  & \textcolor{red}{$\downarrow$39.2\%} \\
         \hline
        ReDet & 0.097 & 0.065 & 0.102 \\
        GOOD (ReDet) & \fs 0.052 & \fs 0.039 & \fs 0.046  \\
        ~ & \textcolor{red}{$\downarrow$46.3\%} & \textcolor{red}{$\downarrow$40.0\%}  & \textcolor{red}{$\downarrow$54.9\%} \\
        \hline
	\end{tabular}
	\label{ablationangle}
\end{table}

\subsubsection{On Each Component}

Table~\ref{ablation1}~studies the impact of each component in GOOD, namely, style hallucination, HRoI consistency learning ($\mathcal{L}_{HCL}$ in Eq.~\ref{HRoICL}), 
rotation-aware consistency learning ($\mathcal{L}_{RCL}$ in Eq.~\ref{RRoICL}) and style consistency learning ($\mathcal{L}_{SEC}$ in Eq.~\ref{JSDloss}). 

It can be seen that, only implementing style hallucination slightly improves the generalization, yielding 0.99\%, 0.91\% and 0.57\% mAP gain on the DOTA-V1.0, DOTA-V1.5 and DOTA-V2.0 unseen target domain, respectively.
Implementing consistency learning on both HRoIs ($\mathcal{L}_{HCL}$) and RRoIs ($\mathcal{L}_{RCL}$) can improve the generalization ability on unseen target domains.
Specifically, $\mathcal{L}_{HCL}$ yields a 0.07\%, 0.61\% and 0.46\% mAP improvement on DOTA-V1.0, DOTA-V1.5 and DOTA-V2.0 unseen target domain. 
Similarly, $\mathcal{L}_{RCL}$ yields a 0.30\%, 0.72\% and 0.41\% mAP improvement on DOTA-V1.0, DOTA-V1.5 and DOTA-V2.0 unseen target domain. 
Finally, the constraint on style hallucination ($\mathcal{L}_{SEC}$) can further improve the generalization, yielding an mAP improvement of 1.11\%, 1.05\% and 1.39\% on the three unseen target domains, respectively. 

\subsubsection{Feature Space Analysis} 

To better understand how the proposed GOOD alleviates the domain gap for oriented object detection, 
we extract the feature embeddings for samples from each domain, and display them by t-SNE visualization. 
Both the proposed GOOD and its baseline are implemented by the above operations for comparison. 
Ideally, the feature embedding of a domain generalized oriented object detector is supposed to demonstrate no distribution shift for samples from different domains.

Fig.~\ref{tsne}~(a) and (b) compare the feature embedding distribution of the RoI-Trans baseline and the proposed GOOD.
Similarly, Fig.~\ref{tsne}~(a) and (b) compare the feature embedding distribution of the ReDet baseline and the proposed GOOD.
Both result comparisons show that the proposed GOOD allows the samples from the DOTA-V1 source domain and the unseen SODA, FAIR1M and HRSC target domains to be more uniformly clustered than the baseline, which indicates its better generalization ability on unseen target domains.

\subsubsection{On Rotation Angle Prediction}

To investigate if the proposed GOOD is able to improve the rotation angle prediction on unseen target domains, we report the root-mean-square deviation (RMSD) between the rotation angle prediction and the rotation angle ground truth, under the SODA$\rightarrow$\{DOTA-1.0, DOTA-1.5, DOTA-2.0\} setting.
The reported RMSD is an average of the RMSD value of 
each object from a certain domain. 
Ideally, a more domain generalized oriented object detector is supposed to infer more precise rotation angle prediction on unseen target domains than the baseline. 

For more comprehensive inspection, two baselines, namely, RoI-Tran and ReDet are both involved.
Table.~\ref{ablationangle}~shows that when the proposed GOOD is embedded into either RoI-Tran or ReDet, the RMSD (in rad) of the rotation angle can be significantly reduced.
These outcomes show the proposed GOOD can improve the rotation angle prediction on unseen target domains. 

\subsubsection{On Style from Each Scale}

The proposed GOOD implements the style hallucination on the image encoder features of each scale.
How the CLIP contributes to the feature hallucination of each scale deserves especial investigation. 

Table~\ref{ablation2}~systematically studies the impact of style hallucination on the feature representation from each scale, denoted as $\mathbf{F}^1$, $\mathbf{F}^2$, $\mathbf{F}^3$ and $\mathbf{F}^4$.
Specifically, we start from only hallucinating the first block feature $\mathbf{F}^1$, and add the block feature one by one.
It is observed that, the style hallucination on each scale positively contributes to the detector's generalization ability.
Besides, the more scale features are hallucinated, the better generalization ability an oriented object detector shows. 

\begin{table}[!t]
    \centering
        \footnotesize
        \caption{Ablation study on style augmentation. Experimental setting: SODA$\rightarrow$\{DOTA-V1.0, -V1.5, V2.0\}. Evaluation metric $mAP_{50}$ presented in percentage.
        }
	\begin{tabular}{cccc|ccc}
		\hline
		\multicolumn{4}{c|}{Style Hallucination} & \multicolumn{3}{c}{Target Domains} \\ 
        \hline
        $\mathbf{F}^1$ & $\mathbf{F}^2$ & $\mathbf{F}^3$ & $\mathbf{F}^4$ & D-1.0 & D-1.5 & D-2.0 \\
        \hline
        $\checkmark$ & & & & 43.25 & 46.12 & 42.53 \\
        $\checkmark$ & $\checkmark$ & & & \rd 43.65 & \rd 46.38 & \rd 43.04 \\ 
        $\checkmark$ & $\checkmark$ & $\checkmark$ & & \nd 43.96 & \nd 46.83 & \nd 43.56 \\
        $\checkmark$ & $\checkmark$ & $\checkmark$ & $\checkmark$ & \fs 44.65 & \fs 47.54 & \fs 43.88 \\
        \hline
	\end{tabular}
	\label{ablation2}
\end{table}

\subsubsection{On Different Distance Metric}

The proposed style consistency learning ($\mathcal{L}_{SEC}$) poses the content consistency between the pre- and post- hallucinated features from an aerial image, which is implemented by minimizing the JSD distance metric.
It is necessary to investigate if some other distance metrics achieve a similar effect, or whether the JSD distance can show the optimal performance on unseen target domains. 

\begin{figure*}[!t]
  \centering
   \includegraphics[width=1.0\textwidth]{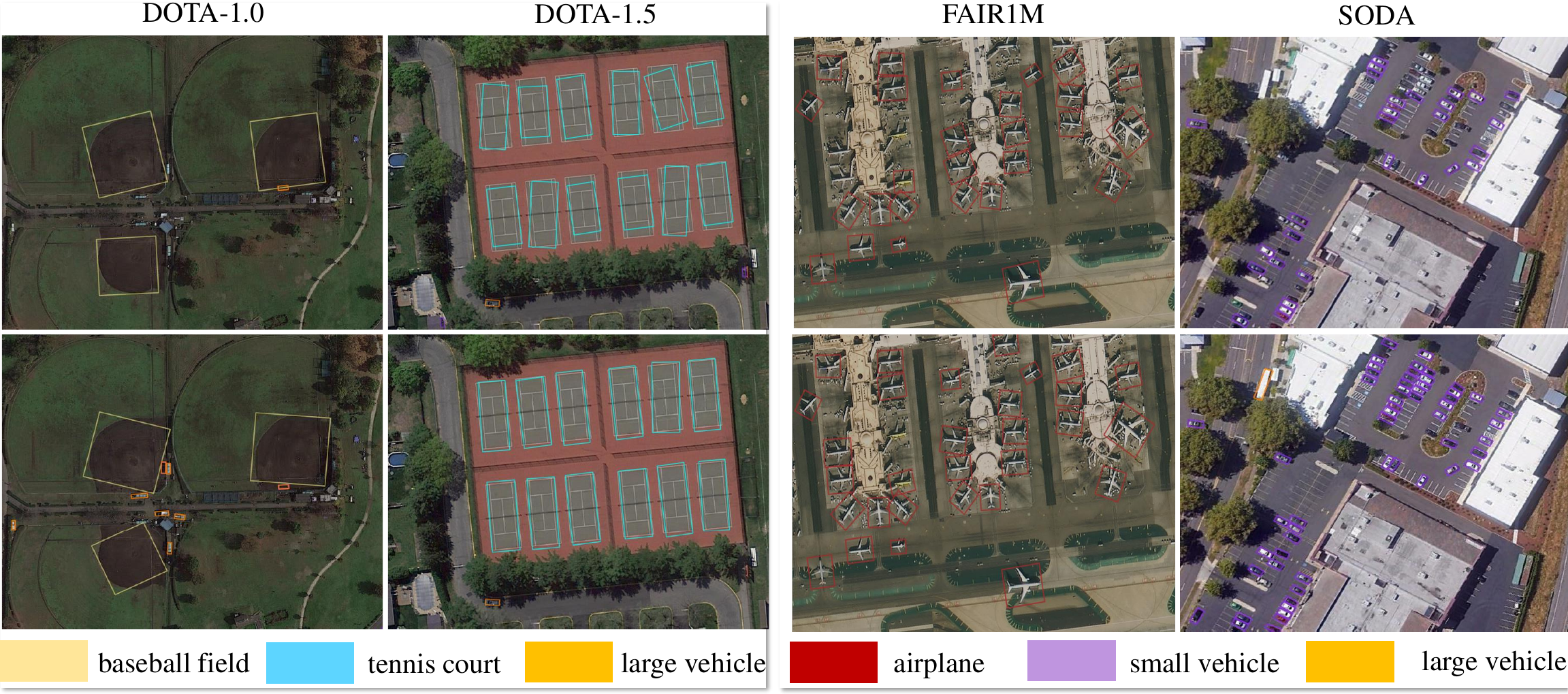} 
\caption{Visualization analysis of the predicted results. First row: baseline; Second row: the proposed GOOD. First and second column: FAIR1M $\rightarrow$ \{DOTA-1.0, DOTA-1.5\}; third and forth column: DOTA-1.0 $\rightarrow$ \{FAIR1M, SODA\}. Compared with baseline, the proposed GOOD not only reduces the omission, but also provides more price orientation prediction. Zoom in for better view.}
   \label{vispred}
\end{figure*}

\begin{figure*}[!t]
  \centering
   \includegraphics[width=1.0\textwidth]{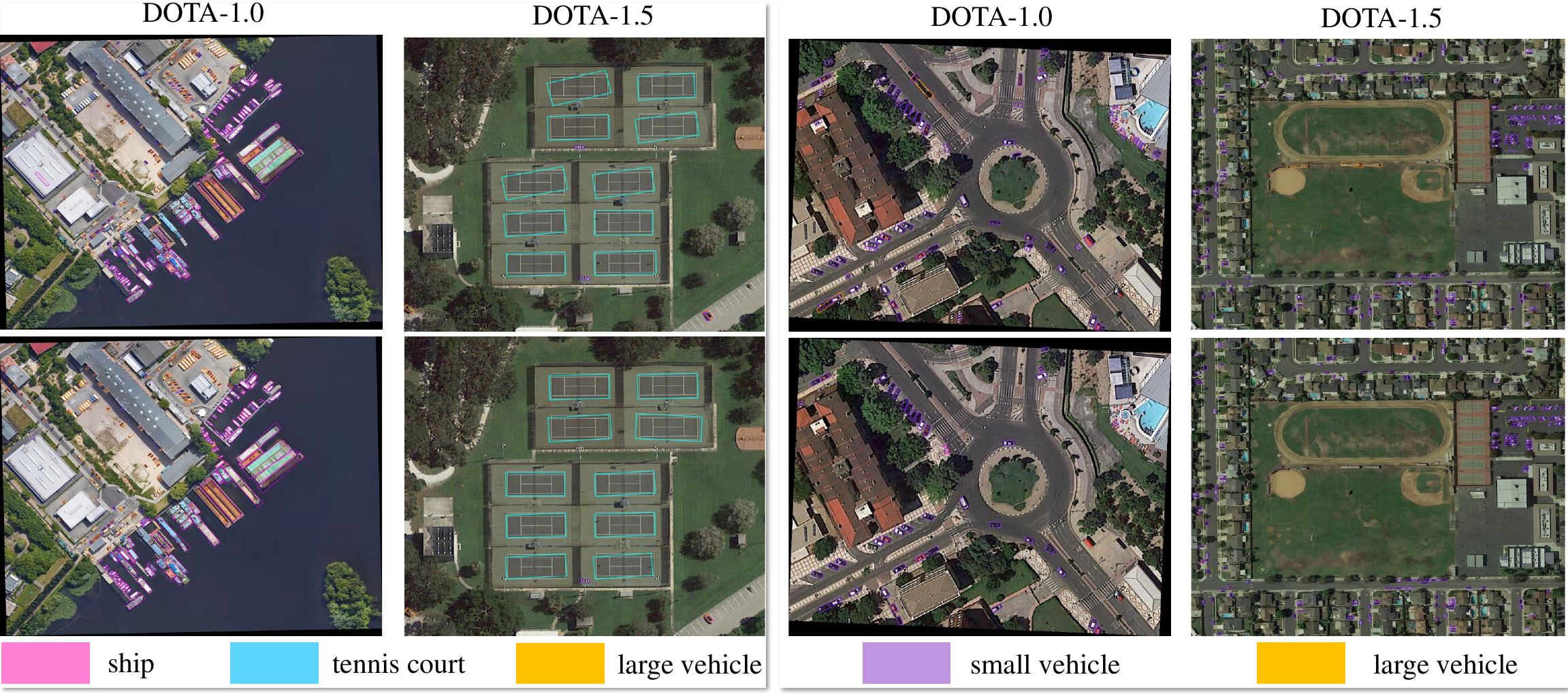} 
\caption{
Visualization analysis of the predicted results under the setting of SODA $\rightarrow$ \{DOTA-1.0, DOTA-1.5, DOTA-2.0\}. 
Compared with baseline, the proposed GOOD not only reduces the omission, but also provides more precise orientation prediction. Zoom in for better view. }
   \label{vispred2}
\end{figure*}

To this end, Table~\ref{ablation3}~investigates the impact of different distance metric on $\mathcal{L}_{SEC}$ (in Eq.~\ref{JSDloss}).
The JSD distance metric in $\mathcal{L}_{SEC}$ is compared with two commonly-used distance metrics, namely, $l$-2 distance and K-L distance.
It leads to an mAP improvement of 1.06\% and 0.65\% on the SODA and FAIR1M unseen target domains than the $l$-2 distance metric.
Similarly, it leads to an mAP improvement of 0.67\% and 0.15\% on both unseen target domains than the K-L distance metric.

To conclude, the style consistency learning from $l$-2 and K-L distance is less significant than the JSD distance.
As discussed in Sec.~\ref{sec3.4}, JSD distance is more capable to handle scenarios when the category scores of pre- and post- hallucinated samples vary greatly in terms of the numerical value, thus yielding better generalization performance.

\begin{table}[!t]
    \centering
        \footnotesize
        \caption{Impact on the distance metric in Eq.~\ref{JSDloss}. Experimental setting: DOTA$\rightarrow$\{SODA, FAIR1M\}. Evaluation metric $mAP_{50}$ presented in percentage.
        }
	\begin{tabular}{c|ccc}
		\hline
		\multirow{2}{*}{Distance Metric} & \multicolumn{2}{c}{Target Domains} \\ 
        \cline{2-3}
         ~ & SODA & FAIR1M \\
        \hline
        $l$-2 loss & 56.13 &  45.56 \\
         K-L loss & 56.52 & 46.07  \\
         JSD loss & \fs 57.19 & \fs 46.21 \\
        \hline
	\end{tabular}
	\label{ablation3}
\end{table}

\subsection{Visualization}

To study if the proposed GOOD is able to visually predict more reasonable and more precise rotated bounding boxes on multiple object categories from these unseen target domains, we display some prediction results on large-scale aerial images.
Both the results from the ReDet \cite{han2021redet} baseline and the proposed GOOD are displayed, which we denote as Baseline and Ours, respectively.

\paragraph{FAIR1M $\rightarrow$ \{DOTA-1.0, -1.5, -2.0\}}

The first two columns in Fig.~\ref{vispred}~provide the visual results on large-scale aerial images.
In the first scene, compared with the baseline, the proposed GOOD predicts more precise orientation for \textit{baseball field}. 
In the second scene, compared with the baseline, the proposed GOOD predicts more precise orientation for \textit{tennis court}.
Besides, in both scenes, more \textit{large vehicle} objects are detected by GOOD.

\paragraph{DOTA $\rightarrow$ \{SODA, FAIR1M\}}
The second two columns in Fig.~\ref{vispred}~show some visual predictions when the baseline and the proposed GOOD is generalized to the FAIR1M and SODA target domain, respectively.
The proposed GOOD not only shows more precise orientation predictions on airplanes, but also more capable to discern the vehicle category. 

\paragraph{SODA $\rightarrow$ \{DOTA-1.0, -1.5, -2.0\}}
Fig.~\ref{vispred2}~provides the visual results on this cross-domain setting.
In the third road scene, compared with the baseline, the proposed GOOD significantly reduces the false alarms of small vehicles (upper left of the image) and large vehicles (at the bottom of the image).
In the fourth urban scene, compared with the baseline, the proposed GOOD reduces some false alarms of \textit{small vehicles} along the road and \textit{large vehicles} at the center of the image.

\section{Conclusion}
\label{sec5}

In this paper, we make an early exploration to learn domain generalized oriented object detection from a single source domain. 
It is particularly challenging as the detector need to represent not only stable content but also precise orientation under the cross-domain style variation. 
In the proposed GOOD, the CLIP-driven style hallucination is constrained by the proposed rotation-aware content consistency learning and style-consistency learning, which allows the model to learn generalizable rotation and content representation.
Experiments on a series of domain generalization settings show that the proposed GOOD has significantly stronger generalization than existing state-of-the-art oriented object detectors.

\paragraph{Limitation \& Future Work.}
One may argue the proposed GOOD only follows the two-stage paradigm. But compared with the one-stage paradigm, 
it provides a more precise description between orientation and region of interests.
It also provides more access to enrich the style and constrain the content consistency from shallow to deep.
On the other hand, we focus on existing domain generalized detection settings, where the cross-domain object categories are consistent.
In the future, we will advance GOOD to open-world scenarios, which can further discover unknown categories from different domains.

\section*{\centering Acknowledgment}
This work was supported by the National Natural Science Foundation of China under contracts No.U22B2011 and No.62325111.

\bibliographystyle{cas-model2-names}

\bibliography{cas-refs}







\end{document}